\documentclass[twoside,leqno,twocolumn]{article}
\usepackage{ltexpprt}

\usepackage{amssymb}
\usepackage{amsmath}

\usepackage{booktabs} % For formal tables
\usepackage{float}

\usepackage{tikz}
\usetikzlibrary{positioning}

\usepackage[font=bf]{caption}
\usepackage{subcaption}

\usepackage{url}
\usepackage{enumitem}
\usepackage{pgfplots}

\usepackage{hhline}
\usepackage{multirow}
\usepackage{booktabs}
\usepackage{wrapfig}

\usepackage{hhline}
\usepackage{multirow}
\usepackage{booktabs}
\usepackage{wrapfig}
\usepackage{algorithm}
\usepackage{algpseudocode}

\newtheorem{definition}{Definition}

\begin{document}
\title{Adversarial Extreme Multi-label Classification}

\author{ Rohit Babbar\footnote{Part of this work was done when the author was at MPI for Intelligent Systems} \\
 Aalto University, Helsinki, Finland
  \and
Bernhard Sch\"{o}lkopf \\
 MPI for Intelligent Systems, Tuebingen, Germany
 }
  \date{}
\maketitle
\begin{abstract}
The goal in extreme multi-label classification is to learn a classifier which can assign a small subset of relevant labels to an instance from an extremely large set of target labels. 
Datasets in extreme classification exhibit a long tail of labels which have small number of positive training instances. 
In this work, we pose the learning task in extreme classification with large number of tail-labels as learning in the presence of adversarial perturbations.
This view motivates a robust optimization framework and equivalence to a corresponding regularized objective.

Under the proposed robustness framework, we demonstrate efficacy of Hamming loss function for tail-label detection in extreme classification.
The equivalent regularized objective, in combination with proximal gradient based optimization, performs better than state-of-the-art methods on propensity scored versions of precision@k and nDCG@k(upto 20\% relative improvement over \texttt{PFastreXML} - a leading tree-based approach and 60\% relative improvement over \texttt{SLEEC} - a leading label-embedding approach).
Furthermore, we also highlight the sub-optimality of a sparse solver in a widely used package for large-scale linear classification, which is interesting in its own right.  
We also investigate the spectral properties of label graphs for providing novel insights towards understanding the conditions governing the performance of Hamming loss based one-vs-rest scheme vis-\`{a}-vis label embedding methods.
\end{abstract}
\section{Introduction}
Extreme Multi-label Classification (\textbf{XMC}) refers to supervised learning with a large target label set where each training/test instance is labeled with small subset of relevant labels which are chosen from the large set of target labels.
Machine learning problems consisting of hundreds of thousand labels, are common in various domains such as annotating web-scale encyclopedia \cite{prabhu2014fastxml}, hash-tag suggestion in social media \cite{denton2015user}, and image-classification \cite{deng2010does}.
For instance, all Wikipedia pages are tagged with a small set of relevant labels which are chosen from more than a million possible tags in the collection.
It has been demonstrated that, in addition to automatic labelling, the framework of XMC can be leveraged to effectively address learning problems arising in recommendation systems, ranking and web-advertizing \cite{Agrawal13, prabhu2014fastxml}.
In the context of recommendation systems for example, by learning from similar users' buying patterns in e-stores like Amazon and eBay, this framework can be used to recommend a small subset of relevant items from a large collection in the e-store. 
In the scenarios of ad-display, by learning the browsing behavior of similar users, relevant advertisements can be displayed to a user from an extremely large collection of all possible advertisements.
With applications in a diverse range, designing effective algorithms to solve XMC has become a key challenge for researchers in industry and academia alike.

In addition to large number of target labels, typical datasets in XMC consist of a \textit{similar scale} for the number of instances in the training data and also for the dimensionality of the input space. 
For text datasets, each training instance is a sparse representation of a few hundred non-zero features from the input space having dimensionality of the order hundreds of thousand. 
An an example, a benchmark \textbf{WikiLSHTC-325K} dataset from the Extreme Classification Repository \cite{repo} consists of 1.7 Million training instances which are distributed among 325,000 labels and each training instance sparsely spans a feature space of 1.6 Million dimensions. 
The challenge posed by the sheer scale of number of labels, training instances and features, makes the setup of XMC quite different from that tackled in classical literature in multi-label classification \cite{tsoumakas2009mining}, and hence renders the \textit{direct and off-the-shelf application} of some of the classical methods, such as Random Forests, Decision Trees and SVMs, non-applicable.  
\subsection{Tail Labels} \label{sec:tail}
\begin{figure}[htp]
\scalebox{1.1}{
\centering
\begin{subfigure}[b]{0.45\linewidth}
\includegraphics[width=0.99\textwidth]{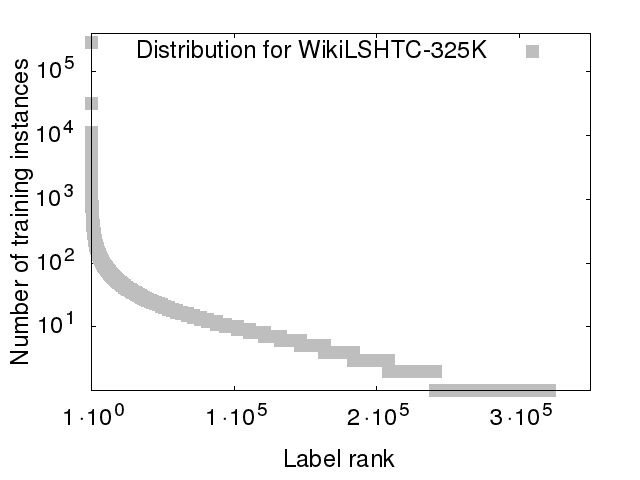}\caption{WikiLSHTC-325K\label{pl:wiki325}}
\end{subfigure}
\begin{subfigure}[b]{0.45\linewidth}
\includegraphics[width=.99\textwidth]{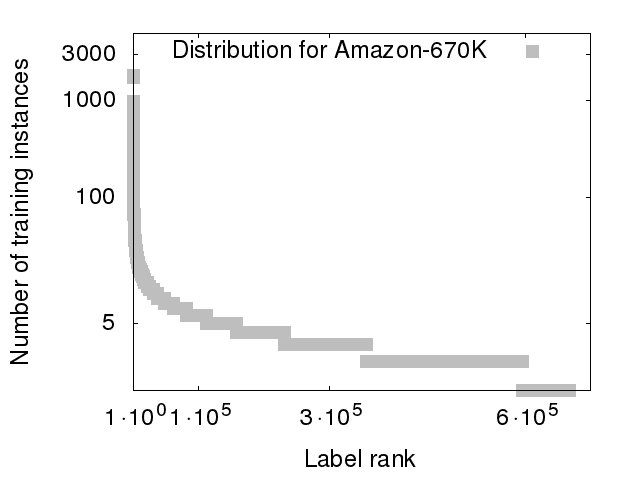}\caption{Amazon-670K\label{pl:amazon670}}
\end{subfigure}
}
\caption{Power-law distribution. Y-axis is on log-scale.}
\label{pl:fig}
\end{figure}

An important statistical characteristic of the datasets in XMC is that a large fraction of labels are tail labels, i.e., those which have very few training instances that belong to them (also referred to as power-law, fat-tailed distribution and Zipf's law). 
This distribution is shown in Figure \ref{pl:fig} for two publicly available benchmark datasets ( \cite{repo}), \textbf{WikiLSHTC-325K} and \textbf{Amazon-670K} datasets, consisting of approximately 325,000 and 670,000 labels respectively. 
For \textbf{Amazon-670K}, only 100,000 out of 670,000 labels have more than 5 training instances in them (Figure \ref{pl:amazon670}). 

Tail labels exhibit diversity of the label space, and also might contain more informative content compared to head or torso labels consisting of large number of training instances.
Indeed, by predicting well the head labels, an algorithm can achieve high accuracy and yet omit most of the tail labels.
Such behavior is not desirable in many real world applications.
For instance, in movie recommendation systems, the head labels correspond to popular blockbusters---most likely, the user  has already watched these.
In contrast, the tail corresponds to less popular yet equally favored films, like independent movies \cite{export:192375}.
These are the movies that the recommendation system should ideally focus on.
A similar discussion applies to search engine development \cite{radlinski2009redundancy} and hashtag recommendation in social networks \cite{denton2015user}.

From a statistical perspective, it has been conjectured in the recent works that Hamming loss is unsuitable for detection of tail-labels in XMC \cite{Jain16, bhatia2015sparse, prabhu2014fastxml}.
On this assumption, a battery of tree-based \cite{Jain16, prabhu2014fastxml, si2017gradient, niculescu-mizil17a, daume2016logarithmic, jernite2016simultaneous, jasinska} and label embedding methods \cite{bhatia2015sparse,yu2014large,xu2016robust,tagami2017annexml} have been proposed in the literature.
In particular, the work in \cite{Jain16} proposes new loss functions which are sensitive towards the tail-labels by weighing them higher than the head/torso labels.
In this work, we concretely evalutate the efficacy of Hamming loss for tail-label detection XMC.
Concretely, our contributions in this work are the following : \\
\noindent (I) \textbf{Statistically}, we model XMC as learning in the presence of adversarial perturbations. This novel perspective stems from the observation that there is a significant variation in the feature composition of instances in the test set as compared to the training set. 
We thus frame the learning problem as a \textit{robust optimization objective} which accounts for this feature variation by considering adversarial perturbations $\tilde{\textbf{x}}_i$ for each input training instance $\textbf{x}_i$. Allowing worst case perturbations in the form of $||\tilde{\textbf{x}}_i||_\infty$ leads to an equivalent 1-norm regularized objective function.\\
(II) \textbf{Algorithmically}, by exploiting a distributed computing architecture for concurrent training of labels, we design a forward-backward proximal gradient algorithm to minimize the 1-norm regularized objective with a convex upper bound on Hamming loss as the choice for loss function. 
Our investigation also shows that the corresponding solver in the LibLinear package ("-s 5" option) yields sub-optimal solutions because of severe under-fitting. 
Due to its widespread usage in machine learning packages such as scikit-learn, this finding is significant in its own right.\\
(III) \textbf{Empirically}, our results have two major findings. Firstly, contrary to the recent conjectures, we show that our Hamming-loss based algorithm gives state-of-the-art results on benchmark datasets in XMC. For WikiLSHTC-325K dataset, we show 20\% relative improvement over \texttt{PFastreXML} - a leading tree based approach, and 60\% over \texttt{SLEEC} - a leading label embedding method.
Secondly, we demonstrate the statistical strength of 1-norm regularization over 2-norm, for tail-label detection in XMC. In our opinion, this has been unknown to the community perhaps due to sub-optimality of the LibLinear solver.\\
(IV) \textbf{Analytically}, by drawing connections to spectral properties of label graph, we also present novel insights to explain the conditions under which Hamming loss might be suited for XMC vis-\`{a}-vis label embedding methods.
We show that the algebraic connectivity of label graph can be used to explain the variation in the relative performance of various methods as it varies from small datasets consisting of few hundred labels to the extreme regime consisting of hundreds of thousand labels. 

\begin{table*}
\begin{subtable}[t]{0.50\linewidth}
\centering
\scalebox{0.60}{
\begin{tabular}{|c|l|}
\hline
  \textbf{Training}  & \\
    \textbf{instances}  & \\
\hline
 1. & Vision computational investigation into the human representation\\
& and processing of visual information david marr late of the massachusetts\\
   & institute of technology was the author of many seminal articles on\\
   & visual information processing and artificial intelligence.\\
 \hline
 2.  & Foundations of vision it has much to offer everyone who wonders how\\
  & this most remarkable of all senses works karen de valois science.\\
\hline
\textbf{Test} &  \\
\textbf{instance} &  \\
\hline
  & Vision science photons to phenomenology this is monumental work\\
  1.  & covering wide range of topics findings and recent approaches on the\\
  & frontiers anne princeton university stephen palmer is professor of\\
 & psychology and director of the institute of cognitive studies at the\\
 & university of california berkeley \\
\hline
\end{tabular}
}
\caption{\footnotesize Training and Test instances for Label 28503}
\label{tbl:ex1}
\end{subtable}\hfill
\begin{subtable}[t]{0.50\linewidth}
\centering
\scalebox{0.60}{
\begin{tabular}{|c|l|}
%\begin{tabular}{p{2.4cm}|p{2.8cm}|p{2.2cm}|p{1.4cm}|p{1.2cm}}
\hline
  \textbf{Training}  & \\
  \textbf{instances}  & \\
\hline
 1. & Manhunt in the african jungle vhs 1943 an american secret agent  \\ 
 & matches wits with nazi agents in casablanca contains 15 episodes \\
 \hline
 2.  & Men vs black dragon vhs 1943 fifteen episodes of 1942 serial showing \\
 & government agents as they exposed the infamous black dragon society \\
 &  an axis spy ring intent on crippling the war effort \\
\hline
\textbf{Test instance}   \\
\hline
  1.  &  And the vhs 1942 this is classic movie from 1942 with action \\
  & death defying stunts and breathless cliffhangers \\
\hline
\end{tabular}
}
\caption{\footnotesize Training and Test instances for Label 246910}
\label{tbl:ex2}
\end{subtable}
\caption{\footnotesize Training and test instance for two tail labels from \textbf{Amazon-670K} dataset depicting variation from training to test set instances.}
\label{tbl:examples}
\end{table*}
Furthermore, in contrast to an earlier robustness-based approach \cite{xu2016robust} in XMC, which models tail-labels as outliers in a label embedding framework, our work models data scarcity of tail-labels in XMC as training in the presence of adversarial perturbations and hence to robust optimization. 
As we shall present, this perspective also draws connections to recent advances in making deep networks robust to specifically designed perturbations to real images by training them on adversarial samples.
Not only does our approach give state-of-the-art results with Hamming loss but also exhibits the statistical strength of 1-norm regularization in tail-label detection.
\section{Problem Setup}\label{sec:setup}
Let the training data, given by $\mathcal{T} = \left\lbrace(\textbf{x}_1,\textbf{y}_1), \ldots ,(\textbf{x}_N,\textbf{y}_N) \right\rbrace$ consist of input feature vectors $\textbf{x}_i \in \mathcal{X} \subseteq \mathbb{R}^D $ and respective output vectors $\textbf{y}_i \in \mathcal{Y} \subseteq \{0,1\}^L$ such that $\textbf{y}_{i_{\ell}}=1$ iff the $\ell$-th label belongs to the training instance $\textbf{x}_i$. 
Recall that in the XMC settings, the cardinality of the set of target labels (denoted by $L$) can be of the order of hundreds of thousand or even millions. 
Similar magnitudes are typical for the training set size ($N$) and feature set dimensionality ($D$) from which each of the training and test instance is sparsely drawn. 
For each label $\ell$, sign vectors $\textbf{s}^{(\ell)} \in \{+1, -1\}^N$ can be constructed, such that $\textbf{s}^{(\ell)}_{i} = +1 $  if and only if $\textbf{y}_{i_\ell} = 1$, and -1 otherwise.

The goal in XMC is to learn a multi-label classifier in the form of a vector-valued output function $f : \mathbb{R}^D \mapsto \{0,1\}^L$. 
As is common in most of the contemporary research in XMC, the training/test instances in our setup are given by a Bag-of-words representation. 
To enable the development of deep learning methods for XMC, raw text corresponding to these datasets has also been recently added to the repository \cite{repo}.

\subsection{Motivation for Adversarial View-point} 
The fat-tailed distribution of training instances among labels implies that most labels have very few training instances that belong to them. 
This scarcity of training instances leads to a significant change in the composition of features in the test set compared to the training set, even though the \textit{underlying distribution} generating the training and test set is same \textit{in principle}. 
For the tail labels, those features which were active in the training set might not appear in test set, and vice-versa. 

This behavior is demonstrated for two of the tail labels extracted from the raw data corresponding to \textbf{Amazon-670K} dataset (provided by the authors of \cite{liu2017deep}). 
The tail label in Table (\ref{tbl:ex1}) corresponds to book titles and editor reviews for books on Computer Vision and Neuroscience, while the label in Table (\ref{tbl:ex2}) provide similar descriptions for VHS Tapes on Action and Adventure genre.
Note that, in both cases, there is a siginificant variation in the features/vocabulary and content from training set to test set instances. 
Even though to a human reader, the semantics of instances in training and test set may be similar, this might not be so obvious to a learning machine due to considerable feature variation. 
Also, for a given test instance, there may be other labels with a similar vocabulary in the training distribution than the true label.
This phenomenon can be viewed as a setup in which an adversary is generating test examples such that the vocabulary of the test set instances is quite different from those in the training set, and significantly so for tail labels.
\subsection{Robust Optimization For Tail-labels} \label{sec:method}
With the above motivation of an inherent adversarial setup in XMC, we appeal to the robust optimization framework which allows for the possibility of perturbations in the training data.
For every training instance $\textbf{x}_i$, we consider a perturbation $\tilde{\textbf{x}}_i \in \mathbb{R}^D$, which will model the feature variation from training to test set. 

We recall Hamming loss function, for predicted output vector $\hat{\textbf{y}}$ and the ground truth label vector $\textbf{y}$, which is defined as  $\ell_H(\textbf{y}, \hat{\textbf{y}}) = \frac{1}{L}\sum_{\ell=1}^{L}I[\textbf{y}_{\ell} \neq \hat{\textbf{y}}_{\ell}]$, where $I[.]$ is the indicator function.
Hamming loss reduces to 0-1 loss over individual labels and hence can be minimized independently over each of them.
For its concrete evaluation for tail-label detection in XMC, we focus on classifier $f$, whose functional form  is composed of $L$ binary  classifiers. 
In other words, the classifier $f$ is parametrized by $\textbf{W} \in \mathbb{R}^{D \times L} := \left[ \textbf{w}^{(1)}, \ldots, \textbf{w}^{(L)}\right]$.

Taking perturbations $\tilde{\textbf{x}}_i$ into account and replacing the 0-1 loss by hinge-loss as its convex upper bound, the weight vector $\textbf{w}^{(\ell)}$ for label $\ell$ with sign vector $\textbf{s}^{(\ell)}$, is learnt by minimizing the following robust optimization objective (without super-script ($\ell$) for clarity)
\begin{displaymath}
\min_{\textbf{w}} \max_{(\tilde{\textbf{x}}_1,\ldots,\tilde{\textbf{x}}_N)}  \sum_{i=1}^N\max[1-\textbf{s}_i (\langle \textbf{w},\textbf{x}_i - \tilde{\textbf{x}}_i \rangle), 0 ] 
\end{displaymath}
The following theorem from \cite{xu2009robustness} shows that if the norm of a perturbation is bounded in the non-regularized robust optimization framework, then it is equivalent to regularizing with the dual norm without considering perturbations in the input.
\begin{theorem}\cite{xu2009robustness}
Let  $\tilde{\textbf{\emph{x}}}_{i} \in \mathbb{R}^{D}$ and $\widetilde{\textbf{X}} := \left\{ (\tilde{\textbf{\emph{x}}}_{1},\ldots,\tilde{\textbf{\emph{x}}}_{N}) | \sum_{i=1}^N||\tilde{\textbf{\emph{x}}}_{i}|| < \lambda' \right\}$.
Assuming non-separability of the training data, the following robust optimization problem 
\begin{equation}\label{eq:robustsvm}
  \min_{\textbf{\emph{w}}} \max_{(\tilde{\textbf{\emph{x}}}_{1},\ldots,\tilde{\textbf{\emph{x}}}_{N}) \in \widetilde{\textbf{X}}}  \sum_{i=1}^N \max[1-\textbf{\emph{s}}_{i} ( \langle \textbf{\emph{w}},\textbf{\emph{x}}_i - \tilde{\textbf{\emph{x}}}_{i} \rangle), 0 ]
\end{equation}
is equivalent to regularized but non-robust optimization problem
\begin{equation}\label{eq:l1regl1svm}
\min_{\textbf{\emph{w}}}  \lambda' ||\textbf{\emph{w}}||_* + \sum_{i=1}^N \max[1-\textbf{\emph{s}}_{i} ( \langle \textbf{\emph{w}},\textbf{\emph{x}}_i \rangle), 0 ]
\end{equation}
where $||.||_*$ is the dual norm of $||.||$.
\end{theorem}
\noindent{\textbf{Choice of Norm}}
From the above theorem, the choice of norm in the bound on the perturbations in the formulation in Equation (\ref{eq:robustsvm}) determines the regularizer in equivalent frormulation in Equation (\ref{eq:l1regl1svm}). As shown in Table \ref{tbl:examples}, there can be a \textit{significant variation} in the features distribution from the training set to test set instances. 
We therefore consider the worst case perturbations in the input, i.e., $||.||_\infty$ norm. This is given by $||\tilde{\textbf{x}}_{i}||_\infty := \max_{d=1\ldots D}|\tilde{\textbf{x}}_{i_d}|$. 
It may be noted that changing the input $\textbf{x}$ by small pertubations along each dimension such that $||\tilde{\textbf{x}}||_\infty < \lambda'$ even for small value of $\lambda'$ can change the inner product evaluation $\textbf{w}^T\textbf{x}$ significantly. 
By accounting for such perturbations in the training data, the resulting weight vector is robust to variations especially for tail-labels.

Since the dual of $||.||_\infty$ is $||.||_1$ norm, this leads to the $||\textbf{w}||_1$-norm regularized SVM in the optimization problem, and hence resulting in a sparse solution.
For this chioce of norm, the above theorem also shows the equivalence between robustness and sparsity.

From the optimization perspective however, both $||\textbf{\textbf{w}}||_1$ and the hinge-loss $\max[1-\textbf{\textbf{s}}_{i} ( \langle \textbf{\textbf{w}},\textbf{x}_i \rangle), 0 ]$ are non-smooth. 
In the following theorem, we prove that one can replace hinge loss by its squared version given by $(\max[1-\textbf{\textbf{s}}_{i} ( \langle \textbf{\textbf{w}},\textbf{\textbf{x}}_i \rangle), 0 ])^2$ for a different choice of the regularization parameter $\lambda$ instead of $\lambda'$. 
The statistically equivalent problem results in objective function in Equation (\ref{eq:l1regsvm}), which is easier to solve from an optimization perspective. 
\begin{theorem}
The following $||.||_1$ norm regularized objective with hinge loss 
\begin{equation}\label{eq:l1regl11svm}
\min_{\textbf{\emph{w}}}  \lambda' ||\textbf{\emph{w}}||_1 + \sum_{i=1}^N \max[1-\textbf{\emph{s}}_{i} ( \langle \textbf{\emph{w}},\textbf{\emph{x}}_i \rangle), 0 ]
\end{equation}
is equivalent, upto a change in the regularization parameter, to the objective function below with squared hinge loss for some choice of $\lambda$
\begin{equation}\label{eq:l1regsvm}
\min_{\textbf{\emph{w}}}  \lambda ||\textbf{\emph{w}}||_1 + \sum_{i=1}^N (\max[1-\textbf{\emph{s}}_{i} ( \langle \textbf{\emph{w}},\textbf{\emph{x}}_i \rangle), 0 ])^2
\end{equation}
\end{theorem}
The proof technique is similar for regression with Lasso \cite{xu2010robust}, and derived here for classification with hinge loss.
Before proceeding to the proof, we present a definition of \textit{weak efficieny} of a solution.
\begin{definition}
Let $g(.) : \mathbb{R}^D \mapsto \mathbb{R}$ and $h(.) : \mathbb{R}^D \mapsto \mathbb{R}$ be two functions. Then $\textbf{w}^*$ is called \textit{weakly efficient} if atleast one of the following holds, (i) $\textbf{w}^* \in \arg\min_{\textbf{w} \in \mathbb{R}^D} g(\textbf{w})$, (ii) $\textbf{w}^* \in \arg\min_{\textbf{w} \in \mathbb{R}^D} h(\textbf{w})$, and (iii) $\textbf{w}^*$ is Pareto efficient, which means that $\nexists$ $\textbf{w}'$ such that $g(\textbf{w}') \leq g(\textbf{w}^*)$  and $h(\textbf{w}') \leq h(\textbf{w}^*)$ with atleast one holding with strict inequality.
\end{definition}
\begin{proof}
A standard result from convex analysis states that for convex functions $g(\textbf{w})$ and $h(\textbf{w})$, the set of optimal solutions for the weighted sum, $ \min_{\textbf{w}} (\lambda_1 g(\textbf{w}) + \lambda_2 h(\textbf{w}))$ where $\lambda_1, \lambda_2 \in [0,+\infty)$ and not being zero together, coincides with the set of weakly efficient solutions.

This means that the set of optimal solutions of $ \min_{\textbf{w}}  (\lambda' ||\textbf{w}||_1 + \sum_{i=1}^N \max[1-\textbf{s}_{i} (\langle \textbf{w},\textbf{x}_i \rangle),0]) $, where $\lambda'$ ranges in $[0,+\infty)$ is the set of weakly efficient solution of $ ||\textbf{w}||_1 $ and $ \sum_{i=1}^N \max[1-\textbf{s}_{i} (\langle \textbf{w},\textbf{x}_i \rangle), 0] $. 
On similar lines, the set of optimal solutions of $ \min_{\textbf{w}} ( \lambda ||\textbf{w}||_1 + \sum_{i=1}^N (\max[1-\textbf{s}_{i} (\langle \textbf{w},\textbf{x}_i \rangle), 0 ])^2) $ where $\lambda$ ranges in $[0,+\infty)$ is the set of weakly efficient solution of $ ||\textbf{w}||_1 $ and $ \sum_{i=1}^N (\max[1-\textbf{s}_{i} ( \langle \textbf{w},\textbf{x}_i \rangle), 0 ])^2$. 
Since taking the square for non-negatives is a monotonic function, it implies that these two sets are identical, and hence are two formulations given in Equations (\ref{eq:l1regl11svm}) and (\ref{eq:l1regsvm}) upto change in the regularization parameter.
\end{proof}
\subsection{Adversarial examples in Deep Learning} 
In this section, we take a brief digression to connect our work to recent advances on training with adversarial exmaples in deep learning. In this context, it has been observed that despite having a good generalization performance, a trained neural network is easily fooled by images which are slight perturbations of a real image \cite{szegedy2013intriguing,goodfellow2014explaining,shaham2015understanding,cisse2017parseval}. 
The goal is, therefore, to robustify the predictions of the deep network by automatic generation of artificial images which are \textit{specifically pertubed versions} of real images and training the network on the generated images also. 
It has been shown in \cite{shaham2015understanding}, that the \textit{Fast Gradient Sign Method} \cite{goodfellow2014explaining} for generation of adversarial examples can also be derived by considering bounded $||\tilde{\textbf{x}}||_\infty$ perturbations around the linearized objective function.

Concretely, let $J(\pmb{\theta},\textbf{x},\textbf{y})$ be the objective function for training the deep network with parameters $\pmb{\theta}$. Then, if a first order approximation of the loss is taken around the given training instance $\textbf{x}$ with small perturbation $\tilde{\textbf{x}}$, it is given by 
\begin{displaymath}
J_{\pmb{\theta},\textbf{y}}( \textbf{x}+ \tilde{\textbf{x}}) \approx J_{\pmb{\theta},\textbf{y}}( \textbf{x}) + \langle\nabla J_{\pmb{\theta},\textbf{y}}( \textbf{x}), \tilde{\textbf{x}}\rangle
\end{displaymath} 
where $\nabla J_{\pmb{\theta},\textbf{y}}( \textbf{x})$ is the gradient of loss function w.r.t to input $\textbf{x}$, which is available from back-propagation. 
The perturbation $\tilde{\textbf{x}}$ which maximizes the loss under the constraint $||\tilde{\textbf{x}}||_\infty < \lambda'$ is given by $\tilde{\textbf{x}} = \lambda'\text{sign} (\nabla J_{\textbf{x}}(\pmb{\theta},\textbf{x},\textbf{y}))$. 
Since the gradient information is available during back-propagation, the adversarial perturbations can be efficiently generated.
These connections suggest that it may be possible to address data scarcity for tail-labels by following a similar approach of sample generation for data augmentation.
Adverarial samples have also been shown to be generated for Question-Answering tasks for NLP \cite{jia2017adversarial} in which sentences are added to mislead a deep learning system towards giving a wrong answer and drastically reducing its answering accuracy.
In our current XMC setup, however, the adversarial nature of the problem is inherent due to the scarcity of training instances in the tail-labels, and learnt model needs to be robust to this behavior.
\subsection{Sub-optimality of Liblinear Solver \cite{fan2008liblinear}}
The formulation in Equation (\ref{eq:l1regsvm}) lends itself to easier optimization and an efficient solution has been implemented in the Liblinear package (as -s 5 argument) by solving a Cyclic Co-ordinate Descent (CCD) procedure.
Not only has it been used as a standard method for large-scale linear solvers in machine learning packages such as scikit-learn and Cran LibLineaR, but it has been used to solve L1-regularized sub-problems appearing in XMC algorithms such as \textit{PFastXML} and \textit{SLEEC}.
A natural question to ask is - \textit{why not use this solver directly if the modeling of XMC with the adversarial setting and the resulting optimization problem are indeed correct.}

We applied the CCD based implementation in LibLinear and found that it gives sub-optimal solution. 
In particular, the CCD solution, (i) underfits the training data, and (ii) does not give good generalization performance.
For concreteness, let $\textbf{w}_{CCD} \in \mathbb{R}^D$ be minimizer of the objective function Equation (\ref{eq:l1regsvm}) and $opt_{CCD} \in \mathbb{R}^+$ be the corresponding optimal value of the objective value attained using the CCD solver.
We demonstrate under-fitting by producing a certificate $\textbf{w}_{Prox} \in \mathbb{R}^D$ with the corresponding objective funtion value $opt_{Prox} \in \mathbb{R}^+$ such that $opt_{Prox} < opt_{CCd}$. 
The construction of the certificate of sub-optimality is obtained by following a proximal gradient procedure in the next section.
The inferior generalization performance of Liblinear is shown in Table \ref{tbl:results1}, which among other methods, provides comparison on the test set of the models learnt by CCD and that learnt by proximal gradient procedure.  
For CCD solver, changing the tolerance condition or increasing the number of iterations had no significant impact on training error reduction.
Due to the wide-spread usage of the sparse solver in LibLinear, this finding is interesting its own right. 

\noindent \textbf{Shrinking heursitics} :
We investigated further the possible reasons for sub-optimal solution for the CCD solver in LibLinear. 
It uses shrinking heuristics for reducing the problem size based on some variable/features which become zero during the process of optimization. 
Let $I(\textbf{w})$ and $b_i(\textbf{w})$ respectively denote the indices of training points with non-zero training error and their miss-classification penalty. Formally, these are given by the following :
\begin{displaymath}
I(\textbf{w}) := \{ i | b_i(\textbf{w}) > 0\} \text{ and } b_i(\textbf{w}) := 1 - \textbf{s}_i\textbf{w}^T\textbf{x}_i
\end{displaymath}

Then the squared hinge loss in Equation (\ref{eq:l1regsvm}) and its derivative w.r.t $\textbf{w}$ can be written as, $\mathcal{L}(\textbf{w}) := \sum_{i \in I(\textbf{w})} (b_i(\textbf{w}))^2$ and $\mathcal{L}'(\textbf{w}) := -2\sum_{i \in I(\textbf{w})} s_i\textbf{x}_i(b_i(\textbf{w}))^2$ respectively. 
The optimiality condition along a co-ordinate $\textbf{w}_j$ is obtained by taking gradient of (4) w.r.t $\textbf{w}_j$
\[ \left\{ 
\begin{array}{l l}
  \mathcal{L}'(\textbf{w}_j) + \lambda = 0 & \quad \mbox{if $\textbf{w}_j > 0$}\\
  \mathcal{L}'(\textbf{w}_j) - \lambda = 0 & \quad \mbox{if $\textbf{w}_j < 0$}\\ 
  -\lambda \leq \mathcal{L}'(\textbf{w}_j) \leq \lambda & \quad \mbox{if $\textbf{w}_j  = 0$}\ \end{array} \right. \]
The violation of the optimality condition along $\textbf{w}_j$ is therefore given by: 
\[ v_j = \left\{ 
\begin{array}{l l}
  |\mathcal{L}'(\textbf{w}_j) + \lambda| & \quad \mbox{if $\textbf{w}_j > 0$}\\
  |\mathcal{L}'(\textbf{w}_j) - \lambda| & \quad \mbox{if $\textbf{w}_j < 0$}\\ 
  \max(\mathcal{L}'(\textbf{w}_j) - \lambda, - \lambda - \mathcal{L}'(\textbf{w}_j) , 0) & \quad \mbox{if $\textbf{w}_j  = 0$}\ \end{array} \right. \]
The shrinking heuristic used in CCD procedure is that if at some iteration $\textbf{w}_j=0$, then $\textbf{w}_j$ is removed from the optimization process if $ -\lambda + M \leq \mathcal{L}'(\textbf{w}_j) \leq -\lambda - M $, where $ M :=  \frac{\max_j(v_j \text{ at previous iteration})}{N}$. This shrinking is conjectured upon the assumption that $\textbf{w}_j$ will not become non-zero later. It is not clear if this is a sufficiently appropriate criterion for variable shrinking. 
%Further investigation on this finding is left as a future work.
\subsection{Certificate Construction by Proximal Gradient}
\begin{algorithm}[H]
\begin{algorithmic}[1]
\Require{Binary training data $(\textbf{X}, \textbf{s})$ and initialize $\textbf{w}_0=\textbf{0}$ }
\Ensure{Learnt weight vector  $\textbf{w}_{Prox}$ for each label independently }
\State t=0
\While{not converged}
    \State $\textbf{u}_t  = \textbf{w}_t -  \gamma_t \mathcal L'({\textbf{w}_t})$
    \State  $\textbf{w}_{t+1} =  \arg\min_{\textbf{w}} \left[\frac{\lambda \gamma_t }{2} ||\textbf{w}||_1 + \frac{1}{2}||\textbf{w} - \textbf{u}_t||_2^2\right]$
    \State $t = t+1$
\EndWhile
\State $\textbf{w}_{Prox} = \textbf{w}_t$ ; return $\textbf{w}_{Prox}$
\end{algorithmic}
\caption{Proximal gradient method to optimize objective (\ref{eq:l1regsvm}) for learning $\textbf{w}_{Prox}$ for label $\ell$}
\label{algorithm:alg1}
\end{algorithm}
\begin{table*}
\centering
\scalebox{0.90}{
\begin{tabular}{|c|c|c|c|c|c|c|}
\hline
  \textbf{Dataset} &  \# \textbf{Training} &  \# \textbf{Features} &  \# \textbf{Labels} &   \multicolumn{2}{c|}{\textbf{}} &  \textbf{Algebraic} \\
  \cline{5-6}
  &(N)& (D) & (L) & \textbf{APpL} & \textbf{ALpP} &  \textbf{Connectivity}, $\lambda_2(G)$ \\
\hline
\textbf{Mediamill} & 30,993  & 120 & \textbf{101} &  1902.1 & 4.4 & \textbf{0.46}  \\
\textbf{Bibtex} & 4,880  & 1,836 & \textbf{159} &   111.7 & 2.4 & \textbf{0.30} \\
\textbf{EUR-Lex} & 15,539  & 5,000  & \textbf{3,993} &   25.7 & 5.3 & \textbf{0.22} \\
\textbf{WikiLSHTC-325K} &  1,778,351 &  1,617,899 & \textbf{325,056}  & 17.4 & 3.2 &\textbf{0.002}\\
\textbf{Wiki-500K} & 1,813,391 & 2,381,304 & \textbf{501,070}  & 24.7 & 4.7 & \textbf{0.001} \\
\textbf{Amazon-670K} & 490,499 & 135,909 & \textbf{670,091}  & 3.9 & 5.4 &\textbf{0.0001}\\
\hline
\end{tabular}
}
\caption{\footnotesize Multi-label datasets from XMC repository. APpL and ALpP represent average points per label and average labels per point respectively. Mediamill and Bibtex do not have tail-labels.
The algebraic connectivity is calculated in Section \ref{sec:gt}.
}
\label{tbl:alldata}
\end{table*}
Proximal methods have been effective in addressing large-scale non-smooth convex problems which can be written as sum of a differentiable function with Lipschtiz-continuous gradient and a non-differentiable function.
We use this scheme to construct the certificate $\textbf{w}_{Prox}$ by solving the optimization problem in Equation (\ref{eq:l1regsvm}) using a forward-backward proximal procedure described in Algorithm \ref{algorithm:alg1}. 
The two main steps in the algorithm are given in line 3 and 4. 
Line 3 (called the forward step), where gradient with respect to the differentiable part of the objective is taken, which in this case is $\mathcal{L}(\textbf{w})$.
The step size $\gamma_{t}$, which can be thought as inverse of the Lipshitz constant of $\mathcal{L}'(\textbf{w}_t)$, is estimated for a new weight $\textbf{w}'$ by starting at a high value and decreasing fractionally until \cite{bach2011convex}:
\begin{displaymath}
\mathcal{L}(\textbf{w}') \leq \mathcal{L}(\textbf{w}_t) + \mathcal{L}'(\textbf{w}_t)^T(\textbf{w}' - \textbf{w}_t) + 1/(2\gamma_t)||\textbf{w}' - \textbf{w}_t||_2^2
\end{displaymath}
Line 4 is the backward or proximal step in which minimization problem involving the computation of the proximal operator has a closed-form solution for $||\textbf{w}||_1$.
It given by the \textit{soft-thresholding operator}, which for the $d$-th dimension at the $t$-th iterate is :
\begin{equation}\label{eq:softt}
\textbf{w}_{d_{t+1}} = sign\left(\textbf{u}_{d_t}\right) \max\left(\left(|\textbf{u}_{d_t}| - \lambda\right), 0\right)
\end{equation}
Note the forward-backward procedure detailed in Algorithm \ref{algorithm:alg1} learns the weight vector corresponding to each label. 
Similar to DiSMEC \cite{dismec}, since the computations are independent for each label, it can be invoked in parallel over as many cores as are available for computation to learn $\textbf{W}_{Prox} = \left[ \textbf{w}^{(1)}_{Prox}, \ldots, \textbf{w}^{(L)}_{Prox}\right]$. 
We call our proposed method \texttt{PRoXML} which stands for \textbf{P}arallel \textbf{Ro}bust e\textbf{X}treme \textbf{M}ulti-\textbf{L}abel classification. 
The convergence of the forward-backward scheme for proximal gradient has been studied in \cite{combettes2007douglas}. 

\begin{figure}[htp]
\centering
\scalebox{0.5}{
\includegraphics[width=0.65\textwidth]{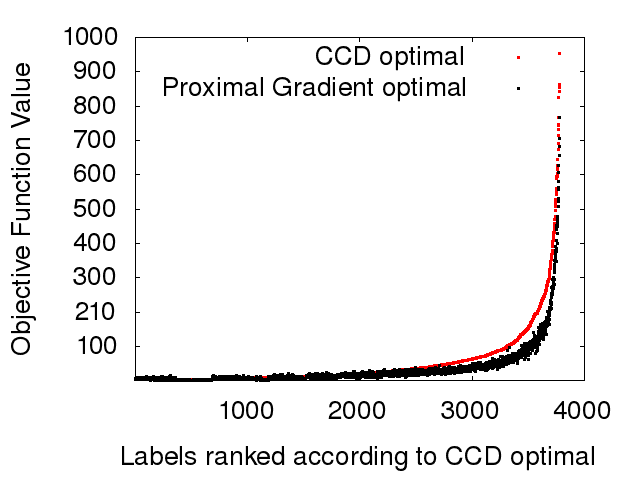}
}
\caption{Comparison of $opt_{Prox}$ and $opt_{CCD}$ over individual labels for \textbf{EUR-Lex} dataset.}
\label{fig:obs}
\end{figure}

Figure \ref{fig:obs} shows the variation in the LibLinear optimization objective for the \textbf{EUR-Lex} dataset between LibLinear CCD and proximal gradient solvers. 
For approximately 90\% of the labels, the objective value obtained by Algorithm \ref{algorithm:alg1}, was lower than that obtained by LibLinear, which in some cases could be as low as half.
It may be noted that LibLinear objective uses miss-classification penalty $C$ instead of the regularization hyper-parameter $\lambda$. 
To enable the comparison, cross-validation was performed for both separately, then the best value $\textbf{W}_{Prox}$ learnt from Algorithm \ref{algorithm:alg1} was substituted to the LibLinear objective function to compute the objective value.
It may be noted that our method does not perform any label embedding, and hence is orthogonal to the embedding-scheme which learns a separate embedding for tail-labels by considering them as outliers in a label embedding space \cite{xu2016robust}.
\section{Experimental Analysis}
\textbf{Dataset description and evaluation metrics} We perform empirical evaluation on publicly available datasets from the XMC repository curated from sources such as Wikipedia and Delicious \cite{mencia2008efficient, mcauley2013hidden}.
The detailed statistics of the datasets are shown in Table \ref{tbl:alldata}.
The datasets exhibit a wide range of properties in terms of number of training instances, features, and labels.
\textbf{MediaMill} and \textbf{Bibtex} datasets are small scale datasets and do not exhibit tail-label behavior.
The last column shows the Algebraic Connectivity of label graph (detailed in Section \ref{sec:gt}), which essentially measures the degree of connectedness of labels based on their co-occurrences in the training data. 
The calculation of algebraic connectivity based on algebraic graph theoretic considerations, is described in Section \ref{sec:gt}.

With applications in recommendation systems, ranking and web-advertizing, the objective of the machine learning system in XMC is to correctly recommend/rank/advertize among the top-k slots.
Propensity scored variants of precision@k and nDCG@k capture prediction accuracy of a learning algorithm at top-k slots of prediction, and also the diversity of prediction by giving higher score for predicting rarely occurring tail-labels. 
For label $\ell$, its propensity $p_\ell$ is related to number of its positive training instances $N_\ell$ by $p_\ell  \propto  1/\left(1+ e^{-\log(N_\ell)}\right)$. With this formulation, $p_\ell \approx 1$ for head-labels and $p_\ell << 1$ for tail-labels. 
Let $\textbf{y} \in \{0,1\}^L$ and $\hat{\textbf{y}} \in \mathbb{R}^L$ denote the true and predicted label vectors respectively.
As detailed in \cite{Jain16}, propensity scored variants of $P@k$ and $nDCG$ are given by
\begin{eqnarray}
PSP@k(\hat{\textbf{y}},\textbf{y}) \hspace{-0.1in} & := & \hspace{-0.1in} \frac{1}{k} \sum_{\ell \in rank_k{(\hat{\textbf{y}})}}{\textbf{y}_\ell}/p_\ell \\
PSnDCG@k(\hat{\textbf{y}},\textbf{y}) \hspace{-0.2in}& := \hspace{-0.2in}& \frac{PSDCG@k}{\sum_{\ell=1}^{\min(k, ||\textbf{y}||_0)}{\frac{1}{\log(\ell+1)}}}
\end{eqnarray}
where $PSDCG@k := \sum_{\ell \in rank_k{(\hat{\textbf{y}})}}{[\frac{\textbf{y}_\ell}{p_\ell\log(\ell+1)}]}$ , and $rank_k(\textbf{y})$ returns the $k$ largest indices of $\textbf{y}$.

To match against the ground truth, as suggested in \cite{Jain16}, we use $100*\mathcal{G}(\{\hat{\textbf{y}}\})/\mathcal{G}(\{\textbf{y}\})$ as the performance metric. For $M$ test samples, $\mathcal{G}(\{\hat{\textbf{y}}\}) = \frac{-1}{M}\sum_{i=1}^{M}\mathcal{L}(\hat{\textbf{y}}_i,\textbf{y})$, where $\mathcal{G}(.)$ and $\mathcal{L}(.,.)$ signify gain and loss respectively. 
The loss $\mathcal{L}(.,.)$ can take two forms, (i)$\mathcal{L}(\hat{\textbf{y}}_i,\textbf{y}) =  - PSP@k$, and (ii) $\mathcal{L}(\hat{\textbf{y}}_i,\textbf{y}) =  - PSnDCG@k$.
This leads to the two metrics which are finally used in our comparison in Table \ref{tbl:results1}, denoted by (P1,P3,P5) and (C1,C3,C5) for k=1,3,5.
The results on vanilla versions of these metrics in which $p_{\ell}=1 \forall \ell$ are shown in Table \ref{tbl:results2}.
\begin{table*}
\label{tbl:results}
\begin{subtable}[t]{0.50\linewidth}
\centering
\scalebox{0.80}{
\begin{tabular}{c|rrrrrr}
  \toprule
  Algorithm & N1(\%) & N3(\%)& N5(\%)& P1(\%) & P3(\%)& P5(\%)\\
  \midrule
  \texttt{SLEEC} & \textbf{70.1} & \textbf{72.3} & \textbf{73.1} & \textbf{70.1} & \textbf{72.7} & \textbf{74.0} \\
  \texttt{LEML} & 66.3 & 65.7 & 64.7 & 66.3 & 65.1 & 63.6 \\ 
  \texttt{FastXML} & 66.6 & 66.0 & 65.2  & 66.6 & 65.4 & 64.3 \\
  \texttt{PFastXML} & 66.8 & 66.5 & 65.6 & 66.8 & 65.9 & 64.7   \\
  \texttt{Rocchio} & - & - & - & - & - & - \\
  \texttt{PFastreXML} & 66.8 & 66.5 & 65.6 & 66.8 & 65.9 & 64.7 \\
 \texttt{Parabel} & 61.2 & 60.2& 59.5 & 63.4 & 62.8 & 62.1   \\  
  \texttt{PD-Sparse} & 62.2 &  61.0 &  57.2 & 62.2 & 59.8 & 54.0 \\ 
  \texttt{CCD-L1} & 63.9 & 62.8 & 62.0  & 63.6 & 60.2 & 59.7 \\
  \texttt{DiSMEC} & 66.5 & 65.5 & 65.2  & 66.5 & 65.1 & 63.7 \\
  \midrule
  \texttt{PRoXML} & 64.3 & 63.6 & 62.8 & 64.3 & 61.3 & 60.8	 \\
  \bottomrule
\end{tabular}
}
\caption{\textbf{MediaMill}, N = 31K, D = 120, L = 101}\label{tbl1:one}

\scalebox{0.80}{
\begin{tabular}{c|rrrrrr}
  \toprule
  Algorithm & N1(\%) & N3(\%)& N5(\%)&  P1(\%)& P3(\%)& P5(\%)\\
  \midrule
  \texttt{SLEEC} & \textbf{51.1}  & \textbf{52.9} &\textbf{56.0} & \textbf{51.1} & \textbf{53.9} & 59.5 \\
  \texttt{LEML} & 47.9 & 50.2 & 53.5 & 47.9 & 51.4 & 57.5 \\ 
  \texttt{FastXML} & 48.5 & 51.1 & 54.3 & 48.5 & 52.3 & 58.8 \\
  \texttt{PFastXML} & 49.7  & 52.3 & 55.6 & 49.7 & 53.5 & 59.6 \\
  \texttt{Rocchio} & - & - & -  & - & - & -  \\
  \texttt{PFastreXML} & 49.7  & 52.3 & 55.6 & 49.7 & 53.5 & \textbf{59.6} \\
    \texttt{Parabel} & 41.2 & 44.8 & 48.8 & 41.2 & 45.8 & 54.5   \\
  \texttt{PD-Sparse} & 48.3 & 48.4 & 50.7 & 48.3 & 48.7 & 52.9 \\ 
  \texttt{CCD-L1} & 49.9 & 51.6 & 54.9  & 49.9 & 52.1 & 57.9 \\
  \texttt{DiSMEC} & 50.2 & 52.0 & 55.7 & 50.2 & 52.2 & 58.6\\
  \midrule
  \texttt{PRoXML} & 50.1 & 52.1 & 55.1 & 50.1 & 52.0 & 58.3\\
  \bottomrule
\end{tabular}
}
\caption{\textbf{Bibtex}, N = 4880, D = 1836, L = 159}\label{tbl1:two}

\scalebox{0.80}{
\begin{tabular}{c|rrrrrr}
  \toprule
  Algorithm & N1(\%) & N3(\%)& N5(\%)& P1(\%)& P3(\%)& P5(\%)\\
  \midrule
  \texttt{SLEEC} & 35.4 & 38.8 & 40.3 & 35.4 &  39.8 & 42.7 \\
  \texttt{LEML} & 24.1 & 26.4 & 27.7 & 24.1 & 27.2 & 29.1 \\ 
  \texttt{FastXML} & 27.6 & 33.2 & 36.2  & 27.6 & 35.3 & 39.9 \\
  \texttt{PFastXML} & 39.9 & 42.2 & 43.2&39.9 &  43.0 & 44.5   \\
   \texttt{Rocchio} & 39.6 & 39.3 & 39.6 & 39.6 & 39.1 & 39.7   \\
  \texttt{PFastreXML} & 43.8 & 45.9 & 46.5 & 43.8 & 46.4 & 47.3 \\
   \texttt{Parabel} & 37.7 & 43.4 & 46.1 & 37.7 & 44.7 & 48.8   \\
  \texttt{PD-Sparse} & 38.2 & 40.9 & 42.8  & 38.2  & 42.7& 44.8 \\ 
  \texttt{CCD-L1} & 37.8 & 40.5 & 42.3  & 37.8 & 41.6 & 44.1 \\
  \texttt{DiSMEC} & 41.2 & 44.3 & 46.9  & 41.2 &  45.4 & 49.3 \\
  \midrule
  \texttt{PRoXML} & \textbf{45.2} & \textbf{47.5} & \textbf{49.1} & \textbf{45.2}& \textbf{48.5} &\textbf{51.0} \\
  \bottomrule
\end{tabular}
}
\caption{\textbf{EUR-Lex}, N = 15K, D = 5K, L = 4K}\label{tbl1:three}

\end{subtable}\hfill
\begin{subtable}[t]{0.50\linewidth}
\centering
\scalebox{0.80}{
\begin{tabular}{c|rrrrrr}
  \toprule
  Algorithm & N1(\%) & N3(\%)& N5(\%)&  P1(\%) &P3(\%)& P5(\%)\\
  \midrule
  \texttt{SLEEC} &20.5  & 22.4 & 23.5  & 20.5 & 23.3 & 25.2 \\
    \texttt{LEML} & 3.4 & 3.6 & 3.9  & 3.4 & 3.7 & 4.2 \\ 
    \texttt{FastXML} & 16.5 & 19.7 & 21.7  & 16.5 &  21.1 & 23.7 \\
  \texttt{PFastXML} & 25.4 & 26.4  & 27.2  & 25.4 & 26.8 & 28.3 \\  
 \texttt{Rocchio} & 30.4 & 29.5  & 29.7  & 30.4 & 29.2 & 30.3 \\  
  \texttt{PFastreXML} & 30.8 & 31.2 & 32.1  & 30.8 & 31.5 & 33.0 \\
  \texttt{Parabel} & 28.7 & 35.2 & 38.1 & 28.7 & 35.0 &  38.6  \\   
   \texttt{PD-Sparse} & 28.3 & 31.9 & 33.6 & 28.3 & 33.5 & 36.6 \\ 
  \texttt{CCD-L1} & 27.8 & 31.6 & 34.3  & 27.8 & 30.6 & 33.9 \\   
  \texttt{DiSMEC} & 29.1 & 35.9 & 39.4 & 29.1 & 35.6 & 39.4 \\
   \midrule
  \texttt{PRoXML} & \textbf{34.8} & \textbf{38.7} & \textbf{41.5} & \textbf{34.8} & \textbf{37.7} & \textbf{41.0} \\
  \bottomrule
\end{tabular}
}
\caption{\textbf{Wiki-325K}, N = 1.78M, D = 1.62MK, L = 325K}

\scalebox{0.80}{
\begin{tabular}{c|rrrrrr}
  \toprule
  Algorithm & N1(\%) & N3(\%)& N5(\%)& P1(\%) &P3(\%)& P5(\%)\\
  \midrule
  \texttt{SLEEC} & 21.1 & 20.9 & 23.1  & 21.1 & 21.0 & 20.8 \\
  \texttt{LEML} & 3.2 & 3.1 & 3.3  & 3.2 & 3.4 & 3.5 \\ 
    \texttt{FastXML} & 22.5 & 21.5 & 22.1 & 22.5 & 21.8 & 22.4 \\
  \texttt{PFastXML} & 22.2 & 21.6 & 21.8 & 22.2 & 21.3 & 21.6 \\
    \texttt{Rocchio} & 29.8 & 28.4 & 28.3 & 29.8 & 27.5 &  27.4 \\
  \texttt{PFastreXML} & 29.2 & 28.7 & 28.3 & 29.2 & 27.6 & 27.7 \\
   \texttt{Parabel} & 28.8 & 31.2 & 35.5 & 28.8 & 31.9 & 34.6  \\
  \texttt{PD-Sparse} & - & - &  - & - & - & - \\ 
  \texttt{CCD-L1} & 29.8 & 30.2 & 32.5  & 29.8 & 30.2 & 33.1 \\
  \texttt{DiSMEC} & 31.2 & 33.7 & 37.1 & 31.2 & 33.4 &  37.0 \\
  \midrule
  \texttt{PRoXML} & \textbf{33.1} & \textbf{35.2} & \textbf{39.0}  & \textbf{33.1}& \textbf{35.0} & \textbf{39.4} \\
  \bottomrule
\end{tabular}
}
\caption{\textbf{Wiki-500K}, N = 181K, D = 238K, L = 500K}\label{tbl1:seven}

\scalebox{0.80}{
\begin{tabular}{c|rrrrrr}
  \toprule
  Algorithm & N1(\%) & N3(\%)& N5(\%)& P1(\%) &P3(\%)& P5(\%)\\
  \midrule
  \texttt{SLEEC} & 20.6 & 22.6 & 24.4 & 20.6 & 23.3 & 26.0 \\
  \texttt{LEML} & 2.0 & 2.2 & 2.3  & 2.0 & 2.2 & 2.4 \\ 
    \texttt{FastXML} & 20.2 & 22.9 & 25.2 & 20.2 & 23.8 & 27.2 \\
  \texttt{PFastXML} & 27.1 & 27.9 & 28.6 & 27.1 & 28.2 & 29.3 \\
   \texttt{Rocchio} & 28.5 & 29.2 & 29.8  & 28.5&  29.4 & 30.3 \\
  \texttt{PFastreXML} & 28.0 & 28.8 & 29.4 & 28.0 & 29.5 & 30.1 \\
   \texttt{Parabel} & 27.6 & 28.4 & 29.9 & 27.6 & 31.0 & 34.1   \\
  \texttt{PD-Sparse} & - & - & -  & - & - & - \\ 
  \texttt{CCD-L1} & 19.4 & 20.2 & 20.8  & 19.4 & 21.1 & 22.7 \\ 
  \texttt{DiSMEC} & 27.8 & 28.8 & 30.7 & 27.8  & 30.6 &  34.2 \\
  \midrule
  \texttt{PRoXML} & \textbf{30.8} & \textbf{31.7} & \textbf{32.6} & \textbf{30.8}  & \textbf{32.8} & \textbf{35.1} \\
  \bottomrule
\end{tabular}
}
\caption{\textbf{Amazon-670K}, N = 490K, D = 136K, L = 670K}\label{tbl1:six}
\end{subtable}

\caption{Propensity Scored nDCG@k (denoted Nk) and Propensity Scored Precision@k (denoted Pk) for k=1,3,5. 
\texttt{PD-Sparse} could not scale \textbf{Wiki-500K} and \textbf{Amazon-670K}, marked as '-'. 
\texttt{Rocchio} refers to \texttt{$Rocchio_{1000}$} in the text, in which Rochhio classifier is run over top 1,000 labels predicted by \texttt{PFastXML}. Since there are no tail-labels in \textbf{Bibtex} and \textbf{MediaMill}, it was not run on these dataset.
}\label{tbl:results1}
\end{table*}
\begin{table*}
\label{tbl:results}
\begin{subtable}[t]{0.50\linewidth}
\centering
\scalebox{0.80}{
\begin{tabular}{c|rrrrrr}
  \toprule
  Algorithm & N1(\%) & N3(\%)& N5(\%)& P1(\%) & P3(\%)& P5(\%)\\
  \midrule
  \texttt{SLEEC} & \textbf{87.8} & \textbf{81.5} & \textbf{79.2} & \textbf{87.2} & \textbf{73.4} & \textbf{59.1} \\
  \texttt{LEML} & 84.0 & 75.2 & 71.9 & 84.0 & 67.2 & 52.8 \\ 
  \texttt{FastXML} & 84.2& 75.4 & 72.3  & 84.2 & 67.3 & 53.0 \\
  \texttt{PFastXML} & 84.1 &  75.6 & 72.4 & 84.1 & 67.7 & 53.2   \\
  \texttt{Rocchio} & - & - & - & - & - & - \\
  \texttt{PFastreXML} &  84.1 &  75.6 & 72.4 & 84.1 & 67.7 & 53.2   \\
 \texttt{Parabel} & 83.4 & 74.4 & 70.9 & 83.4 & 66.3 & 51.7  \\  
  \texttt{PD-Sparse} & 81.8 &  70.2 & 63.7 & 81.8 & 62.5 & 45.1 \\ 
  \texttt{CCD-L1} & 85.8 & 76.4  & 74.7  & 85.8 & 67.4 & 52.5 \\
  \texttt{DiSMEC} & 87.2 & 78.5 & 76.5 & 87.2 & 69.3 & 54.1 \\
  \midrule
  \texttt{PRoXML} & 86.5 & 77.3 & 75.6 & 86.5 & 68.4 & 	53.2 \\
  \bottomrule
\end{tabular}
}
\caption{\textbf{MediaMill}, N = 31K, D = 120, L = 101}\label{tbl1:one}

\scalebox{0.80}{
\begin{tabular}{c|rrrrrr}
  \toprule
  Algorithm & N1(\%) & N3(\%)& N5(\%)&  P1(\%)& P3(\%)& P5(\%)\\
  \midrule
  \texttt{SLEEC} &  \textbf{65.0} & \textbf{60.4} &  \textbf{62.6} & \textbf{65.0} & \textbf{39.6} & 28.8 \\
  \texttt{LEML} & 62.5 & 58.2 & 60.5 & 62.5 & 38.4 & 28.2 \\ 
  \texttt{FastXML} & 63.4 & 59.5 & 61.7 & 63.4 & 39.2 & 28.8 \\
  \texttt{PFastXML} & 62.8  & 60.0 & 62.0 & 62.8 & 39.6 & 28.9 \\
  \texttt{Rocchio} & - & - & -  & - & - & -  \\
  \texttt{PFastreXML} & 62.8  & 60.0 & 62.0 & 62.8 & 39.6 & \textbf{28.9} \\
    \texttt{Parabel} & 64.4 & 59.3 & 61.0 & 64.4 & 38.5 & 27.9 \\
  \texttt{PD-Sparse} & 61.2 & 55.8 & 57.3 & 61.2 & 35.8 & 25.7 \\ 
  \texttt{CCD-L1} & 64.1 & 59.2 & 61.3  & 64.1 & 38.7 & 28.4 \\
  \texttt{DiSMEC} & 64.5 & 59.4 & 61.6 & 64.5 & 39.2 & 28.4\\
  \midrule
  \texttt{PRoXML} & 64.4 & 59.2 & 61.5 & 64.4 & 39.0 & 28.2\\
  \bottomrule
\end{tabular}
}
\caption{\textbf{Bibtex}, N = 4880, D = 1836, L = 159}\label{tbl1:two}

\scalebox{0.80}{
\begin{tabular}{c|rrrrrr}
  \toprule
  Algorithm & N1(\%) & N3(\%)& N5(\%)& P1(\%)& P3(\%)& P5(\%)\\
  \midrule
  \texttt{SLEEC} & 79.2 & 68.1 & 61.6 & 79.2 &  64.3 &  52.3\\
  \texttt{LEML} & 63.4 &53.5 & 48.4 & 63.4 & 50.3  & 41.2 \\ 
  \texttt{FastXML} & 71.3 & 59.9 & 50.3 & 71.3  & 62.8 & 51.0 \\
  \texttt{PFastXML} & 72.1 & 61.2 & 52.3 & 72.1 & 63.1 & 51.8   \\
   \texttt{Rocchio} & 73.7 & 63.2 & 58.7 & 73.7 & 63.8 &  52.1  \\
  \texttt{PFastreXML} & 75.4 & 65.9 & 60.7 & 75.4 & 62.7 & 52.5 \\
   \texttt{Parabel} & 80.6 & 71.8 & 66.1 & 80.6 & 68.5 &  57.3  \\
  \texttt{PD-Sparse} & 76.4 & 64.3 & 58.7  & 76.4  & 60.3 & 49.7 \\ 
  \texttt{CCD-L1} & 80.8 & 71.2 & 64.9  & 80.8 & 67.8 &  55.8 \\
  \texttt{DiSMEC} & 82.4 & 72.5 & 66.7  & 82.4 &  68.5 & 57.7 \\
  \midrule
  \texttt{PRoXML} & \textbf{83.4} & \textbf{74.4} & \textbf{68.2} & \textbf{83.4} & \textbf{70.9} &  \textbf{59.1}\\
  \bottomrule
\end{tabular}
}
\caption{\textbf{EUR-Lex}, N = 15K, D = 5K, L = 4K}\label{tbl1:three}
\end{subtable}\hfill
\begin{subtable}[t]{0.50\linewidth}
\centering
\scalebox{0.80}{
\begin{tabular}{c|rrrrrr}
  \toprule
  Algorithm & N1(\%) & N3(\%)& N5(\%)&  P1(\%) &P3(\%)& P5(\%)\\
  \midrule
  \texttt{SLEEC} & 54.8 & 47.2 & 46.1  & 54.8 & 33.4 & 23.8 \\
    \texttt{LEML} & 19.8 & 14.5 & 13.7 & 19.8 & 11.4 & 8.3 \\ 
    \texttt{FastXML} & 49.7 & 33.1 & 24.4  & 49.7 &  45.2 & 44.7 \\
  \texttt{PFastXML}   & 54.8 & 48.7 & 48.1  & 54.8 & 35.8 & 25.8 \\  
 \texttt{Rocchio}    & 55.2 & 49.3 & 49.0  & 55.2  & 36.0 & 26.4  \\ 
  \texttt{PFastreXML} & 56.0  & 50.3 &  50.0 & 56.0 & 36.7 & 27.0 \\
  \texttt{Parabel} & 64.7 & 58.3 & 58.1   & 64.7 & \textbf{42.9} & \textbf{31.6} \\
   \texttt{PD-Sparse} &  61.2 & 55.0   & 54.6  & 61.2 & 39.4 & 28.7 \\
  \texttt{CCD-L1} & 60.6 & 55.2 & 55.0  & 60.6 & 38.6 & 28.5 \\
  \texttt{DiSMEC} &\textbf{64.9} &\textbf{58.5}  &  \textbf{58.4} & \textbf{64.9 }& 42.7  & 31.5 \\
   \midrule
  \texttt{PRoXML} & 63.8 & 57.4  & 57.1  & 63.6 & 41.5 &  30.8\\
  \bottomrule
\end{tabular}
}
\caption{\textbf{Wiki-325K}, N = 1.78M, D = 1.62MK, L = 325K}
\scalebox{0.80}{
\begin{tabular}{c|rrrrrr}
  \toprule
  Algorithm & N1(\%) & N3(\%)& N5(\%)& P1(\%) &P3(\%)& P5(\%)\\
  \midrule
  \texttt{SLEEC} & 48.2 & 22.6 & 21.4  & 48.2 & 29.4 & 21.2 \\ 
  \texttt{LEML} & 41.2 & 18.7 & 17.1  & 41.2 & 30.1 & 19.8 \\ 
    \texttt{FastXML} & 54.1 & 26.4 & 24.7  & 54.1 & 35.5 & 26.2 \\ 
  \texttt{PFastXML} & 55.8 & 27.2 & 25.1  & 55.8 & 35.9 & 26.9 \\ 
   \texttt{Rocchio} & 56.2 & 28.6 & 26.7  & 56.2 & 39.5 & 27.8 \\ 
  \texttt{PFastreXML} & 59.5 & 30.1 & 28.7  & 59.5 & 40.2 & 30.7 \\ 
   \texttt{Parabel} & 67.8 & 38.5 & 36.3  & 67.8 & 48.3 & 37.5 \\  
  \texttt{PD-Sparse} & - & - & -  & - & - & - \\
  \texttt{CCD-L1} & 65.3 & 36.2 & 34.3  & 65.3 & 46.1 & 35.3 \\ 
  \texttt{DiSMEC} & \textbf{70.2} & \textbf{42.1} & \textbf{40.5} & \textbf{70.2}  & \textbf{50.6} & \textbf{39.7} \\
  \midrule
  \texttt{PRoXML} & 68.8 & 39.1 & 38.0 & 68.8 & 48.9 & 37.9 \\
  \bottomrule
\end{tabular}
}
\caption{\textbf{Wiki-500K}, N = 181K, D = 238K, L = 500K}\label{tbl1:seven}

\scalebox{0.80}{
\begin{tabular}{c|rrrrrr}
  \toprule
  Algorithm & N1(\%) & N3(\%)& N5(\%)& P1(\%) &P3(\%)& P5(\%)\\
  \midrule
  \texttt{SLEEC} & 34.7 & 32.7 & 31.5 & 34.7 & 31.2 & 28.5 \\ 
  \texttt{LEML} & 8.1 & 7.3 & 6.8 & 8.1 & 6.8 & 6.0 \\ 
    \texttt{FastXML} & 36.9 & 33.2 & 30.5 & 36.9 & 35.1 & 32.5\\ 
  \texttt{PFastXML} & 35.3 & 33.6 & 30.8  & 36.3 & 32.4 & 31.0 \\ 
   \texttt{Rocchio} & 36.9 & 34.7 & 32.6  & 36.9 & 33.9 & 31.6 \\ 
  \texttt{PFastreXML} & 37.8 & 35.8 & 33.2  & 37.8 & 34.5 & 31.9 \\ 
   \texttt{Parabel} & 44.0 & 41.5 & 39.8  & 44.0 & 39.4 & 36.0 \\ 
  \texttt{PD-Sparse} & - & - & -  & - & - & - \\ 
  \texttt{CCD-L1} & 39.8 & 36.8 & 35.2 & 39.8 & 34.3 & 30.1 \\ 
  \texttt{DiSMEC} & \textbf{44.7} & \textbf{42.1} & \textbf{40.5} &\textbf{44.7}& \textbf{39.7} & \textbf{36.1} \\
  \midrule
  \texttt{PRoXML} & 43.5 & 41.1 & 39.7 & 43.5 & 38.7 & 35.3 \\
  \bottomrule
\end{tabular}
}
\caption{\textbf{Amazon-670K}, N = 490K, D = 136K, L = 670K}\label{tbl1:six}
\end{subtable}

\caption{Vanilla nDCG@k (denoted Nk) and Vanilla Precision@k (denoted Pk) for k=1,3,5. 
\texttt{PD-Sparse} could not scale \textbf{Wiki-500K} and \textbf{Amazon-670K}, marked as '-'. 
\texttt{Rocchio} refers to \texttt{$Rocchio_{1000}$} in the text, in which Rochhio classifier is run over top 1,000 labels predicted by \texttt{PFastXML}. Since there are no tail-labels in \textbf{Bibtex} and \textbf{MediaMill}, it was not run on these dataset.
}\label{tbl:results2}
\end{table*}
\subsection{Methods for comparison} 
\begin{figure*}[htp]
\centering
\begin{subfigure}[b]{0.45\linewidth}
\includegraphics[width=0.99\textwidth]{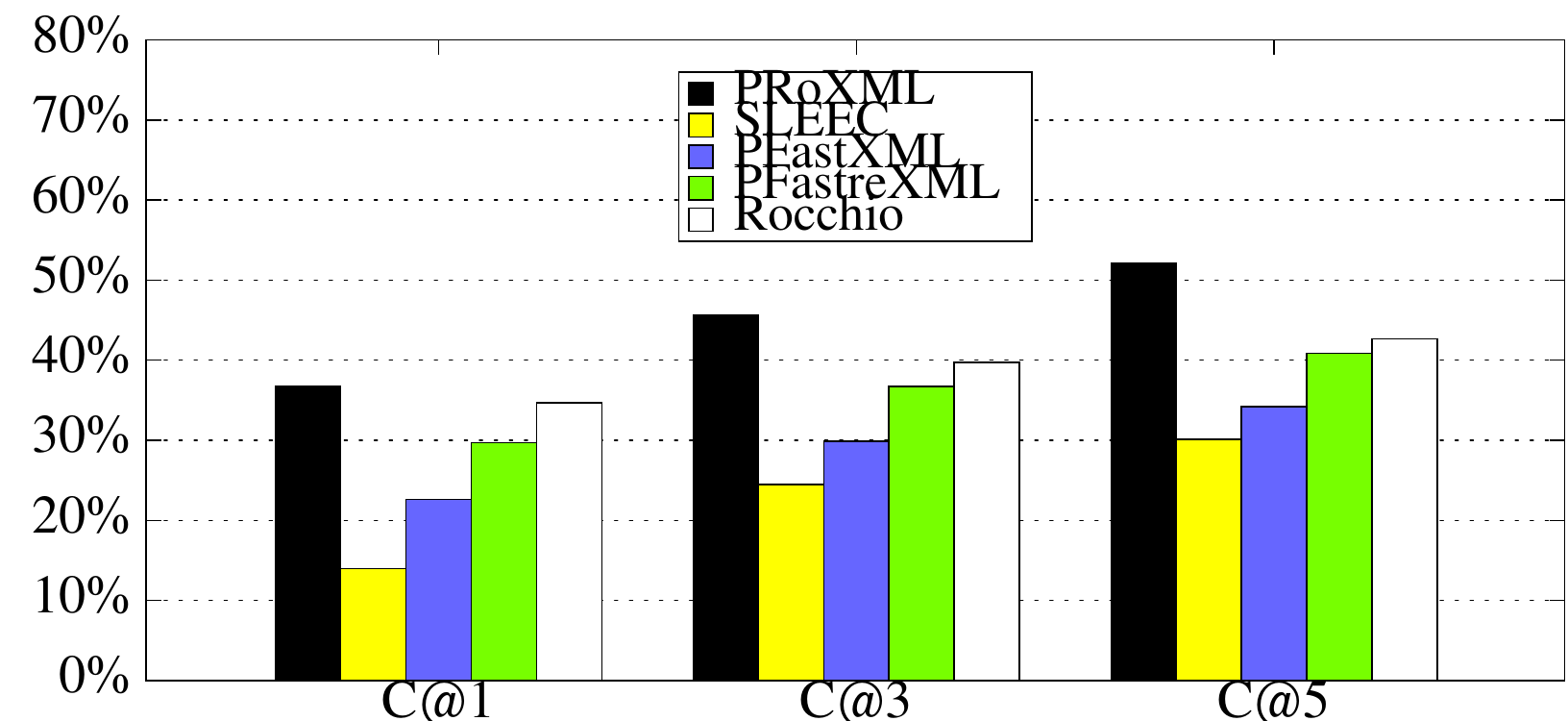}\caption{WikiLSHTC-325K\label{pl:wiki325}}
\end{subfigure}
\begin{subfigure}[b]{0.45\linewidth}
\includegraphics[width=.99\textwidth]{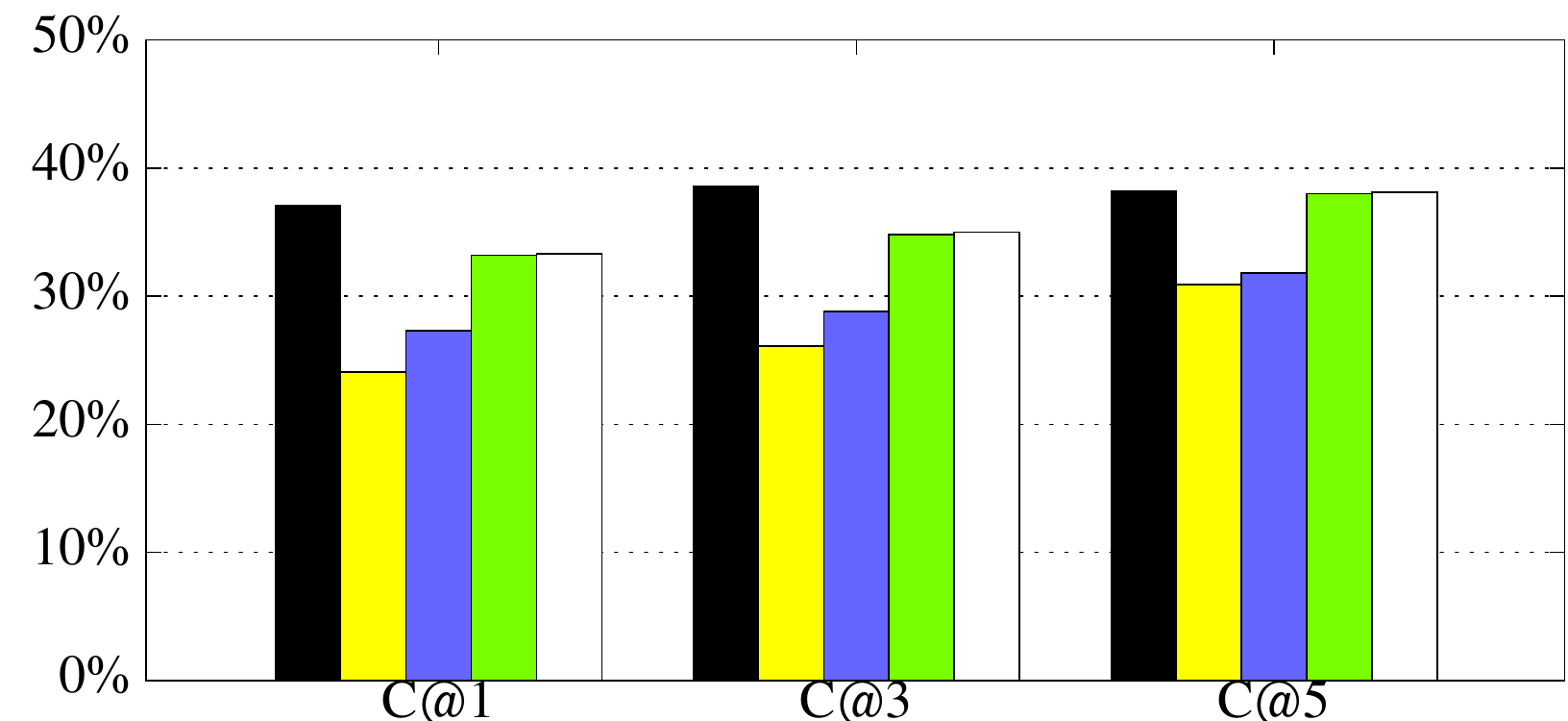}\caption{Amazon-670K\label{pl:amazon670}}
\end{subfigure}
\caption{Label Coverage for various methods on \textbf{WikiLSHTC-325K} and \textbf{Amazon-670K} datasets}
\label{fig:coverage}
\end{figure*}
We compare \texttt{PRoXML} against ten state-of-the-art algorithms:\\
\noindent \textbf{Label Embedding methods} \\
(I) \texttt{SLEEC} \cite{bhatia2015sparse} - It learns sparse \textit{local} embeddings and captures non-linear correlation between the labels.\\
(II) \texttt{LEML} \cite{yu2014large}- It learns a \textit{global} embedding of the label space which may not be suitable when there is a large fraction of tail labels.\\
\noindent \textbf{Tree-based methods} \\
(I) \texttt{PFastXML} \cite{Jain16} - This method optimizes propensity scored metrics and partitions the feature space for faster prediction. \\
(II) \texttt{PFastreXML} \cite{Jain16} - It learns an ensemble of \texttt{PFastXML} and Rocchio classifier applied on the top 1000 labels predicted by \texttt{PFastXML}. 
It is shown to out-perform the production system used in Bing Search (c.f. Section 7 in \cite{Jain16}) and reviewed in detail in Section \ref{sec:related}.\\
(III) \texttt{FastXML} \cite{prabhu2014fastxml} - This is another tree-based method which optimizes vanilla nDCG metric, and is a special case of \texttt{PFastreXML} in which all the propensities are set to 1.\\
(IV) \texttt{Parabel} \cite{Prabhu18b} - This is recently proposed method which learns label partitions by a novel balanced 2-means++ algorithm.\\ \\
\noindent \textbf{Linear methods} \\ 
(I) \texttt{PD-Sparse} \cite{yenpd} - It uses elastic net regularization with multi-class hinge loss and exploits primal and dual sparsity.\\
%and complexity sub-linear in the number of primal and dual variables. 
(II) \texttt{DiSMEC} \cite{dismec} - This is one-vs-rest baseline which achieves state-of-the-art results on vanilla $P@k$ and $nDCG@k$. It minimizes Hamming loss with $l_2$ regularization with weight pruning \textit{heuristic}.\\
(III) \texttt{\texttt{$Rocchio_{1000}$} Classifier} - Referred to as \texttt{Rocchio} in Tables \ref{tbl:results1} and \ref{tbl:results2}, this is obtained by running \texttt{PFastXML} and then using Rocchio classifier for the top 1,000 candidate labels.\\
(IV) \texttt{CCD-L1} - Sparse solver ("-s 5" option) as part of the LibLinear package.

\texttt{PRoXML} was implemented in C++ on 64-bit Linux system using openMP for parallelization.
The code for \texttt{PRoXML} will be made public soon.
For \texttt{PRoXML}, the regularization parameter $\lambda$ was cross-validated for smaller \textbf{MediaMill}, \textbf{Bibtex}, and \textbf{EUR-Lex} datasets and it was fixed to 0.1 for all bigger datasets. 
Due to computational constraints in XMC consisting of hundreds of thousand labels, keeping fixed values for hyper-parameters is quite standard (c.f. Hyper-parameters setting, Section 7 in \cite{Jain16}, and Section 3 in \cite{bhatia2015sparse}, and Section 5 in \cite{prabhu2014fastxml}).
For all other approaches, the results were reproduced as suggested in the papers. 
The relative performance of various methods on propensity scored metrics $PSP@k$ and $PSnDCG@k$ is shown in Table \ref{tbl:results1}, and for vanilla versions is shown in Table \ref{tbl:results2}. The coverage of coverage by taking propensities into account is shown in Figure \ref{fig:coverage}. 
The important observations from these are summarized below : \\
(A) \textbf{For larger datasets} falling in the extreme regime such as \textbf{Amazon-670K}, \textbf{Wiki-500K} and \textbf{WikiLSHTC-325K} which consist of hundreds of thousand labels, \texttt{PRoXML} performs substantially better than both embedding-schemes and tree-based methods such as \texttt{PFastreXML}. For instance, as shown in Table 2(d) for \textbf{WikiLSHTC-325K}, the improvement in $PSP@5$ and $PSnDCG@5$ over \texttt{SLEEC} is almost 60\% and almost 20\% compared to \texttt{PFastreXML}. 
It is important to note that our method works better even on propensity scored metrics than \texttt{PFastreXML} even though its training process is optimizing another metric namely, a convex upper bound on Hamming loss. On the other hand, \texttt{PFastreXML} is minimizing the same metric on which the performance is evaluated.
Due to its robustness properties, \texttt{PRoXML} also performs better on propensity scored metrics than \texttt{DiSMEC} which also minimizes Hamming loss but employs $\ell_2$ regularization followed by weight pruning heuristic for model size reduction.
On the other hand, for vanilla versions of precision@k and nDCG@k, \texttt{DiSMEC} performs better than \texttt{PRoXML}.

These results overall demonstrate the efficacy of Hamming loss in XMC, whether for tail-label detection for propensity-scored metrics or head label detection as in vanilla versions.
In the next section, we will present a spectral graph perspective towards understanding the suitability of Hamming loss based schemes in XMC.
It may be recalled that even mild improvements in large-scale industrial deployments can lead to substantial profits in applications such as recommendations and advertizing.
\\
(B) \textbf{For smaller datasets} such as \textbf{Mediamill} and \textbf{Bibtex} consisting of 101 and 159 labels respectively, embedding based methods \texttt{SLEEC} and \texttt{LEML} perform better or at par with Hamming loss minimizing methods. 
As explained in Section \ref{sec:gt}, this is due to high algebraic connectivity of label graphs in smaller datasets, leading to high correlation between labels. 
This behavior is in stark contrast to datasets in the extreme regime such as \textbf{WikiLSHTC-325K} and \textbf{Amazon-670K} in which Hamming loss minimizing methods significantly outperform label-embedding methods. 
The above differences observed in the performance of small-scale problems vis-\`{a}-vis large-scale problems are indeed quite contrary to the remarks in recent works (c.f. abstract of \cite{Jain16}).\\
(C) \textbf{Label Coverage} is shown in Figure \ref{fig:coverage} (denoted by C@1, C@3, and C@5) for \textbf{WikiLSHTC-325K}, i.e. it measures the fraction of correctly predicted unique labels taking propensities into account.
It is clear that \texttt{PRoXML} performs better than state-of-the-art methods in detecting more unique and correct labels.
From Table \ref{tbl:results1} and Figure \ref{fig:coverage}, it may also be noted that \texttt{$Rocchio_{1000}$} classifier does better than \texttt{PFastXML} on most datasets. 
This indicates that the performance of \texttt{PFastreXML} depends heavily on the good performance of \texttt{$Rocchio_{1000}$} classifier, which in turn is learnt from the top labels predicted by \texttt{PFastXML} classifier.
On the other hand, our method despite not having any such ensemble effects, performs better than \texttt{PFastreXML} and its components \texttt{PFastXML} and \texttt{$Rocchio_{1000}$}.
\section{Discussion - What works, what doesn't and Why?} \label{sec:gt}
We now analyze the empirical results shown in the previous section by drawing connections to spectral properties of label graphs, and determine data-dependent conditions under which Hamming loss minimization is more suited compared to label embedding methods and vice-versa. 
This section also sheds light on qualitative differences between data properties when one moves from small-scale to the extreme regime, and why the intuition for small datasets breaks down at large scale. 

\subsection{Algebraic Connectivity of Label Graphs}
For the training data $\mathcal{T} = \{(\textbf{x}_1,\textbf{y}_1), \ldots ,(\textbf{x}_N,\textbf{y}_N) \}$ consisting of input vectors $\textbf{x}_i$ and respective output vectors $\textbf{y}_i $ such that $\textbf{y}_{i_{\ell}}=1$ iff the $\ell$-th label belongs to the training instance $\textbf{x}_i$.
Consider the adjacency matrix $A(G)$ corresponding to the label graph $G$, whose vertex set $V(G)$ is the set of labels in training set, and the edge weights $a_{\ell, \ell^{'}}$ are defined by $a_{\ell, \ell^{'}}  =  \sum_{i=1}^N \left[ \left(\textbf{y}_{i_{\ell}} = 1\right) \wedge \left(\textbf{y}_{i_{\ell^{'}}} = 1\right) \right]$, where $\wedge$ represents the logical and operator.
The edge between labels $\ell$ and $\ell^{'}$ is weighted by the number of times  $\ell$ and $\ell^{'}$ co-occur in the training data. By symmtery, $a_{\ell, \ell^{'}}  = a_{\ell^{'}, \ell}  \forall \ell, \ell{'} \in V(G)$.
Let $d({\ell})$ denote the degree of label $\ell$, where $d({\ell}) = \sum_{\ell^{'} \in V(G)} a_{\ell, \ell^{'}}$, and $D(G)$ be the diagonal degree matrix $d_{\ell, \ell} = d({\ell})$.
The entries of normalized Laplacian matrix, $L(G)$ is given by :
\[L_{\ell,\ell'} = \left\{ 
\begin{array}{l l}
  1-\frac{a_{\ell, \ell^{'}}}{d_{\ell}} & \quad \mbox{if $\ell=\ell'$ and $d_{\ell} \neq 0$}\\
  -\frac{a_{\ell, \ell^{'}}}{\sqrt{d_{\ell} d_{\ell'}} } & \quad \mbox{if $\ell$ and $\ell'$ are adjacent}\\ 
  0 & \quad \mbox{otherwise}\ \end{array} \right. \]
Let $\lambda_1(G), \ldots , \lambda_L(G)$ be the eigen-values of $L(G)$. A result from spectral graph theory states that $\lambda_2(G) \leq \nu(G) \leq \eta(G)$, where $\nu(G)$ and $\eta(G)$ are respectively the vertex and edge connectivity of $G$. i.e. minimum of vertices and edges to be removed from $G$ to make it disconnected \cite{chung1997spectral}.
Being a lower bound on  $\nu(G)$ and $\eta(G)$, $\lambda_2(G)$ gives an estimate on the connectivity of the label graph. The higher the algebraic connectivity, the more densely connected the labels are in the graph G. 
The last column of Table \ref{tbl:alldata} shows algebraic connectivity for the normalized Laplacian matrix for various datasets. 
Higher values of algebraic connectivity, indicating high degree of connecivity and correlation between labels, are observed for smaller datasets such as \textbf{MediaMill} which consist of only a few hundreds labels.
Lower value is observed for datasets in the extreme regime such as \textbf{WikiLSHTC-325K}, \textbf{WikiLSHTC-500K} and \textbf{Amazon-670K}. 
As opposed to the un-normalized version, the normalized Laplacian is not impacted by the size of the graph. 

\noindent \textbf{Why Hamming loss works for Extreme Classification?}\\ 
Contrary to the assertions in \cite{Jain16}, Hamming loss minimizing one-vs-rest or binary relevance classifier, which trains an independent classifier for every label, works well on datasets in the extreme regime such as \textbf{WikiLSHTC-325K} and \textbf{Amazon-670K}.
In this regime, there is very little correlation between labels that could potentially be exploited in the first place. 
The extremely weak correlation is indicated by crucial statistics shown in Table \ref{tbl:alldata}, which include : lower value of the algebraic connectivity of the label graph $\lambda_2(G)$, fat-tailed distribution of instances among labels and lower values of average number of labels per instance.
The virtual non-existence of correlation indicates that the presence/absence of a given label does not really imply the presence/absence of other labels.
It may be noted that there may be semantic similarity between labels, but there is not enough data, especially for tail-labels, to support that. 
This inherent separation in label graph for larger datasets leads to better performance of one-vs-rest scheme.

\noindent \textbf{Why Label-embedding is suitable for small datasets ?} \\
For smaller datasets that consist of only a few hundred labels (such as \textbf{MediaMill}) and relatively large value for average number of labels per instance, the labels tend to co-occur more often than for datasets in extreme regime. 
In this situation, label correlation is much higher that can be easily exploited by label-embedding approaches leading to better performance compared to one-vs-rest approach. 
This scale of datasets, as is common in traditional machine learning, has been marked by the success of label-embedding methods.
Therefore, it may be noted that conclusions drawn on this scale of problems, such as on the applicability of learning algorithms or suitability of loss functions for a given problem, may not necessarily apply to datasets in XMC.
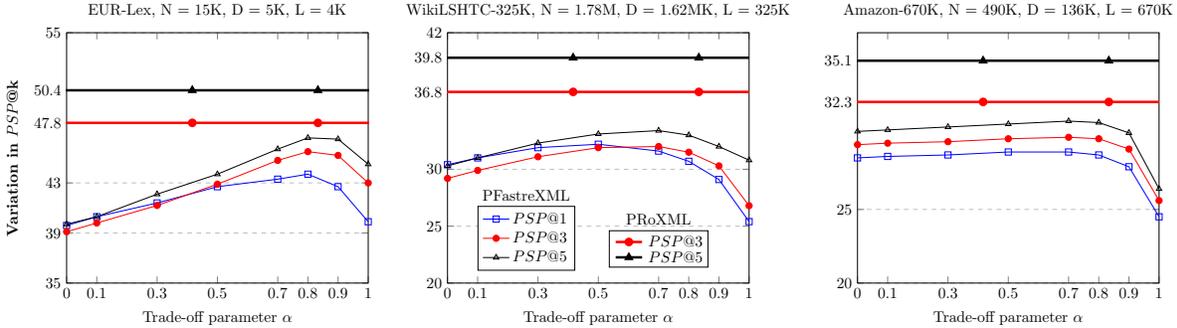
\begin{figure*}[ht]
\scalebox{0.90}{
\centering
\newcommand*{\AddPlotAsymptoteaa}{\addplot[smooth, ultra thick, mark=square, blue] {44.3} }%
	\newcommand*{\AddPlotAsymptotea}{\addplot[smooth, ultra thick, mark=*, red] {47.8} }%
	\newcommand*{\AddPlotAsymptoteb}{\addplot[smooth, ultra thick, mark=triangle, black] {50.4} }%
	
	\newcommand*{\AddPlotAsymptoteca}{\addplot[smooth, ultra thick, mark=square, blue] {32.4} }
	\newcommand*{\AddPlotAsymptotec}{\addplot[smooth, ultra thick, mark=*, red] {36.8} }%
	\newcommand*{\AddPlotAsymptoted}{\addplot[smooth, ultra thick, mark=triangle, black] {39.8} }
	
	\newcommand*{\AddPlotAsymptoteea}{\addplot[smooth, ultra thick, mark=square, blue] {30.3} }
	\newcommand*{\AddPlotAsymptotee}{\addplot[smooth, ultra thick, mark=*, red] {32.3} }%
	\newcommand*{\AddPlotAsymptotef}{\addplot[smooth, ultra thick, mark=triangle, black] {35.1} }

	\begin{tikzpicture}[scale=0.65]
\begin{axis}[
    title={EUR-Lex, N = 15K, D = 5K, L = 4K},
    xlabel={Trade-off parameter $\alpha$},
    ylabel={\textbf{Variation in $PSP$@k}},
    xmin=0.0, xmax=1.0,
    ymin=35, ymax=55,
    xtick={0.0,0.1,0.3,0.5,0.7,0.8,0.9,1.0},
    ytick={35,39,43,55},
     extra y ticks={47.8,50.4},
     extra y tick style={draw=none, inner sep=0pt, outer sep=0.3333em, fill=white, text opacity=1},
    legend style={at={(0.1,0.0)}, anchor=south west},
    ymajorgrids=true,
    grid style=dashed,
]

\addplot[
    color=blue,
    mark=square,
    ]
    coordinates {
    (0,39.6)(0.1,40.3)(0.3,41.4)(0.5,42.7)(0.7,43.3)(0.8,43.7)(0.9,42.7)(1.0,39.9)
    };
    \label{plot_one}
    
    \addplot[
    color=red,
    mark=*,
    ]
    coordinates {
    (0,39.1)(0.1,39.8)(0.3,41.2)(0.5,42.9)(0.7,44.8)(0.8,45.5)(0.9,45.2)(1.0,43.0)
    };
\label{plot_two}

    \addplot[
    color=black,
    mark=triangle,
    ]
    coordinates {
    (0,39.7)(0.1,40.3)(0.3,42.1)(0.5,43.7)(0.7,45.7)(0.8,46.6)(0.9,46.5)(1.0,44.5)
    };
    \label{plot_three}
%    \AddPlotAsymptoteaa;
 \AddPlotAsymptotea;
 \AddPlotAsymptoteb;
\end{axis}
\end{tikzpicture}
~
	\begin{tikzpicture}[scale=0.65]
\begin{axis}[
    title={WikiLSHTC-325K, N = 1.78M, D = 1.62MK, L = 325K},
    xlabel={Trade-off parameter $\alpha$},
    xmin=0.0, xmax=1.0,
    ymin=20, ymax=42,
    extra y ticks       = 0.8,
    xtick={0.0,0.1,0.3,0.5,0.7,0.8,0.9, 1.0},
    ytick={20,25,30,42},
     extra y ticks={36.8,39.8},
legend style={at={(0.1,0.05)}, anchor=south west, label={PFastreXML},},
    ymajorgrids=true,
    grid style=dashed,
]

\addplot[
    color=blue,
    mark=square,
    ]
    coordinates {
    (0,30.4)(0.1,31.0)(0.3,31.9)(0.5,32.2)(0.7,31.6)(0.8,30.7)(0.9,29.1)(1.0,25.4)
    };
    \label{plot_one}
    
    \addplot[
    color=red,
    mark=*,
    ]
    coordinates {
    (0,29.2)(0.1,29.9)(0.3,31.1)(0.5,31.9)(0.7,32.0)(0.8,31.5)(0.9,30.3)(1.0,26.8)
    };
    \label{plot_two}
    
    \addplot[
    color=black,
    mark=triangle,
    ]
    coordinates {
    (0,30.3)(0.1,31.0)(0.3,32.3)(0.5,33.1)(0.7,33.4)(0.8,33.0)(0.9,32.0)(1.0,30.8)
    };
     \label{plot_three}
%     \AddPlotAsymptoteca; \label{plot_four}
    \AddPlotAsymptotec; \label{plot_four}
    \AddPlotAsymptoted ; \label{plot_five} 
\addlegendimage{/pgfplots/refstyle=plot_one}\addlegendentry{$PSP$@1}
\addlegendimage{/pgfplots/refstyle=plot_two}\addlegendentry{$PSP$@3}
\addlegendimage{/pgfplots/refstyle=plot_three}\addlegendentry{$PSP$@5}

%\addlegendimage{/pgfplots/refstyle=plot_four}\addlegendentry{$PSP$@3}
%\addlegendimage{/pgfplots/refstyle=plot_five}\addlegendentry{$PSP$@5}
    
\node [label={PRoXML}, draw] at (axis cs: 0.7,23) {\shortstack[l]{ \ref{plot_four} $PSP$@3 \\ \ref{plot_five} $PSP$@5 }};
\end{axis}

\end{tikzpicture}
    ~
	\begin{tikzpicture}[scale=0.65]
\begin{axis}[
   title={Amazon-670K, N = 490K, D = 136K, L = 670K},
    xlabel={Trade-off parameter $\alpha$},
    xmin=0.0, xmax=1.0,
    ymin=20, ymax=37,
    xtick={0.0,0.1,0.3,0.5,0.7,0.8,0.9, 1.0},
    ytick={20,25,3},
    extra y ticks={32.3,35.1},
    legend pos=south west,
    ymajorgrids=true,
    grid style=dashed,
]

\addplot[
    color=blue,
    mark=square,
    ]
    coordinates {
    (0,28.5)(0.1,28.6)(0.3,28.7)(0.5,28.9)(0.7,28.9)(0.8,28.7)(0.9,27.9)(1.0,24.5)
    };

    \addplot[
    color=red,
    mark=*,
    ]
    coordinates {
    (0,29.4)(0.1,29.5)(0.3,29.6)(0.5,29.8)(0.7,29.9)(0.8,29.8)(0.9,29.1)(1.0,25.6)
    };

    \addplot[
    color=black,
    mark=triangle,
    ]
    coordinates {
    (0,30.3)(0.1,30.4)(0.3,30.6)(0.5,30.8)(0.7,31.0)(0.8,30.9)(0.9,30.2)(1.0,26.4)
    };
%    \AddPlotAsymptotea;
    \AddPlotAsymptotee;
    \AddPlotAsymptotef;
 
\end{axis}
\end{tikzpicture}
}
\caption{ Variation of $PSP$@k  with the trade-off parameter $\alpha$ for (i) \textbf{EUR-Lex}, (ii) \textbf{WikiLSHTC-325K}, and (iii) \textbf{Amazon-670K} datasets. For \texttt{PFastreXML}, $\alpha$ = 0.8. On the left, ($\alpha=0$), represents $\texttt{Rocchio}_{1,000}$ classifier, and on the right ($\alpha=1$), represents \texttt{PFastXML} classifier without re-ranking step. \texttt{PRoXML} works better than \texttt{PFastreXML} for all ranges of $\alpha$ for $PSP$@3 and $PSP$@5. 
$PSP$@1 is not shown for clarity, and it is 44.3, 32.4, and 30.3 respectively.
}
\end{figure*}\label{pfastfig}
\noindent \textbf{What about $PSP@k$ and $PSPnDG@k$ ?} \\
Though $PSP@k$ and $PSPnDG@k$ are appropriate for performance evaluation, these may not right metrics to optimize over during training. 
For instance, if a training instance has fifteen positive labels and we are optimizing $PSP@5$, then as soon as it has correctly classified five out of the fifteen labels correctly, the training process will stop trying to change the decision hyper-plane for this training instance. 
As a result, the information regarding the remaining ten labels is not captured while optimizing the $PSP@5$ metric.
It is possible that at test time, we get a similar instance which has some or all the remaining ten labels which were \textit{not optimized} during training. 
On the other hand, one-vs-rest which minimizes Hamming loss would try to independently align the hyper-planes for all the fifteen labels until these are separated from the rest.
Overall, the model learnt by optimizing is richer compared to that learnt by optimizing $PSP@k$ and $PSPnDG@k$.
Therefore, it leads to better performance on $P@k$ and $nDG@k$ as well as $PSP@k$ and $PSPnDG@k$, when regularized properly.
\subsection{Model Size, and Training/Prediction time}
Due to the sparsity inducing 1-norm regularization, the obtained models are quite sparse and light-weight.
For instance, the model learnt by \texttt{PRoXML} is 3GB in size for \textbf{WikiLSHTC-325K}, compared to 30 GB for \texttt{PFastreXML} on this dataset.
\texttt{PRoXML} proposed in Algorithm 1 uses a distributed training framework thereby exploiting any number of cores as are available for computation. 
The training can be done offline on a distributed/cloud based system for large datasets such as \textbf{WikiLSHTC-325K} and \textbf{Amazon-670K}. 
Faster convergence can be achieved by other methods such as sub-sampling negative examples or warm-starting the optimization with the weights learnt by \texttt{DiSMEC} algorithm to warm-start for faster convergence, via better initialization instead of initializing with an all-zeros solution.
The main aim of this work, however, was to fully explore the statistical properties of Hamming loss and 1-norm regularization obtained by following an adversarial learning framework in the context of XMC.

Prediction speed is more critical for most applications of XMC which demand low latency in domains such as recommendation systems and web-advertizing.
The compact model learnt by \texttt{PRoXML} can be easily evaluated for prediction on streaming test instances. 
This is further aided by distributed model storage which can exploit the parallel architecture for prediction, and takes 2 milliseconds per test instance on average which is thrice as fast as \texttt{SLEEC}, 1,200 times faster than \texttt{LEML} and at par with tree-based methods.
\subsection{Predictive Performance}
\section{Related Work}\label{sec:related}

To handle the large scale of labels in XMC, most methods have focused on two of the main strands, (i) Tree-based methods \cite{Jain16, prabhu2014fastxml,si2017gradient, niculescu-mizil17a, daume2016logarithmic, jernite2016simultaneous, jasinska}, and Label-embedding based methods \cite{bhatia2015sparse,yu2014large,xu2016robust,tagami2017annexml}. Recently, there has been interest in developing distributed linear methods \cite{dismec,yenpd} which can exploit distributed hardware. From a probabilistic view-point, bayesian approaches for multi-label classification have been developed in recent works such as \cite{jain2017scalable, gaure2017probabilistic} and Labeled LDA \cite{papanikolaou2017subset}.

For multi-class classification, the theory of extreme classifcation has been developed in the recent work \cite{yunwenTheory}. In similar context, the behavior of tail-labels for flat and classification with taxonomies has been studied in the previous work \cite{babbar2013flat, babbar2014power, babbar2014re, babbar2016, babbar2016learning}.
%Bayesian methods \cite{jain2017scalable,}
Due to space constraints, we only discuss \texttt{PFastreXML} in detail since it is specifically designed for tail-labels.
\subsection{\textbf{PFastreXML} \cite{Jain16}}
\texttt{PFastreXML} is a state-of-the-art tree-based method which outperformed a highly specialized production system for Bing search engine consisting of ensemble of a battery of ranking methods (cf. Section 7 in \cite{Jain16}).
Learning the \texttt{PFastreXML} classifier primarily involves learning two components, (i) \texttt{PFastXML} classifier - which is an ensemble of trees which minimize propensity scored loss functions, and (ii) a re-ranker which attempts to recover the tail labels missed by \texttt{PFastXML}.
The re-ranker is essentially Rocchio classifier, also called the nearest centroid classifier (Equation 7, Section 6.2 in \cite{Jain16}), which assigns the test instance to the label with closest centroid among the top 1,000 labels predicted by \texttt{PFastXML}. 
%We refer this simply as Rocchio classifier. 
The final score $s_{\ell}$ assigned to label $\ell$ for test instance $\textbf{x}$ is given by a convex combination of scores \texttt{PFastXML} and the Rocchio classifier for top 1,000 label (Equation 8, Section 6.2 in \cite{Jain16}) as follows:
\begin{displaymath}
s_l = \alpha \log P_{PFast} (\textbf{y}_l=1|\textbf{x}) + (1-\alpha) \log P_{Roc_{1,000}}(\textbf{y}_l=1|\textbf{x})
\end{displaymath}
%Setting $\alpha$ to 0.8. 1 and 0 gives \texttt{PFastreXML}, \texttt{PFastXML} and \texttt{$Rocchio_{1,000}$} classifiers respectively.
For \texttt{PFastreXML}, $\alpha$ is fixed to 0.8; setting $\alpha=1$ gives the scores from \texttt{PFastXML} classifier only and $\alpha=0$ gives the scores from \texttt{$Rocchio_{1,000}$} classifier only. 
It may be recalled that, akin to \texttt{FastXML}, \texttt{PFastXML} is also an ensemble of a number of trees, which is typically set to 50.
%In addition to the poorer performance compared to our proposed algorithm, below are some of the short-comings of \texttt{PFastreXML} :\\
Some of its shortcomings in addition to the relatively poorer performance compared to \texttt{PRoXML} are :\\
(I) \textbf{Standalone \texttt{PFastXML}} - Figure 3 shows the variation of $PSP@k$ of \texttt{PFastreXML} with change in $\alpha$ which includes the two extremes (\texttt{PFastXML}, $\alpha = 1$) and (\texttt{$Rocchio_{1,000}$} classifier, $\alpha = 0$) on three datasets from Table \ref{tbl:alldata}. 
Clearly, the performance of \texttt{PFastreXML} depends heavily on good performance of \texttt{$Rocchio_{1,000}$} classifier. 
It may be recalled that one of the main goals of propensity based metrics and \texttt{PFastXML} was better coverage of tail labels. However, \texttt{PFastXML} itself needs to be supported by the additional \texttt{$Rocchio_{1,000}$} classifier for better tail label coverage. 
To the contrary, our method does not need additional such auxiliary classifier.\\
(II) \textbf{Need for Propensity estimation from Meta-data} - To estimate propensities $p_\ell$ using $p_\ell  :=  1/\left(1+C e^{-A\log(N_\ell+B)}\right)$, one needs to compute parameters $A$ and $B$ from some meta-information of the data-source such as Wikipedia or Amazon taxonomies. 
Furthermore, it might not even be possible on some datasets to have an auxillary information, in which case the authors in \cite{Jain16} set it to average of Wikipedia and Amazon datasets, which is quite ad-hoc. Our method does not need propensities for training and hence is also applicable to other metrics for tail-label coverage.\\ 
(III) \textbf{Large Model sizes} - \texttt{PFastreXML} leads to large model size such as 30GB (for 50 trees) for \textbf{WikiLSHTC-325K} data, and 70GB (for 20 trees) for \textbf{Wiki-500K}. Such large model sizes can be difficult to evaluate for making real-time predictions in recommendation systems and web-advertizing. For larger datasets such as \textbf{WikiLSHTC-325K}, the model sizes learnt by \texttt{PRoXML} is around 3GB which is an order of magnitude smaller than \texttt{PFastreXML}.\\
(IV) \textbf{Lots of Hyper-parameters} - \texttt{PFastreXML} has around half a dozen hyper-parameters such as $\alpha$, number of trees in ensemble, and number of instances in the leaf node etc. Also, there is no reason apriori to fix $\alpha=0.8$ even though it gives better generalization performance as shown in Figure 4. To the contrary, our method has just one hyper-parameter which is the regularization parameter.
\section{Conclusion}
We presented the problem with large number tail-labels in XMC framework as learning in the presence of adversarial perturbations, which motivates an equivalent regularized objective function. On benchmark datasets, our proximal gradient procedure to solve the 1-norm regularized objective with Hamming loss outperforms state-of-the-art methods. To provide insights into the observations, we explain the performance gain of one-vs-rest scheme vis-\`{a}-vis label embedding methods. We hope that connections to deep learning for generating samples via adversarial perturbations, opens new research avenues for augmenting data-scarce tail-labels.
\vspace{-0.1in}
\bibliographystyle{abbrv}
\vspace{-0.2in}
\bibliography{LSHTC-biblio1} 
\end{document}